%% file: main.tex
\documentclass{amii}

\input{compact-title.tex}
\input{math_commands.tex}

\usepackage{booktabs}
\usepackage{nicefrac}
\usepackage{longtable}
\usepackage{standalone}
\usepackage{tikz}
\usepackage{xr-hyper}
\usetikzlibrary{calc, positioning, shapes.geometric, arrows.meta, backgrounds,
                decorations.pathreplacing}
\usepackage{hyperref}
\usepackage{url}
\usepackage{wrapfig}
\usepackage{subcaption}
\usepackage{graphicx}
\usepackage{xcolor}
\usepackage[normalem]{ulem} 
\usepackage[outline]{contour}

\contourlength{0.7pt}          
\ExplSyntaxOn
\NewDocumentCommand{\gapuline}{m}
 {
  \seq_set_split:Nnn \l_tmpa_seq { ~ } { #1 }
  \seq_pop_left:NN \l_tmpa_seq \l_tmpa_tl
  \uline{\phantom{\l_tmpa_tl}}\llap{\contour{white}{\l_tmpa_tl}}
  \seq_map_inline:Nn \l_tmpa_seq
   {
    \space \uline{\phantom{##1}}\llap{\contour{white}{##1}}
   }
 }
\ExplSyntaxOff

\newcommand{\customfootnote}[2][\triangle]{%
  \begingroup
  \renewcommand{\thefootnote}{\ensuremath{#1}}%
  \footnote{#2}%
  \addtocounter{footnote}{-1}%
  \endgroup
}

\title{Mastering Atari 2600 Games\\
with Discovered Options}

\author[1,2]{Erik M.~Lintunen}
\author[1,2,3]{Marlos C. Machado}

\shortauthors{Lintunen and Machado}
\shorttitle{Mastering Atari 2600 Games with Discovered Options}

\affiliation[1]{University of Alberta}
\affiliation[2]{Alberta Machine Intelligence Institute (Amii)}
\affiliation[3]{Canada CIFAR AI Chair}

\abstract{Temporal abstractions, often instantiated as options, have long been regarded as a mechanism for accelerating credit assignment, facilitating exploration, and enabling generalisation in reinforcement learning~(RL).
However, developing general option discovery methods that are effective in large-scale, high-dimensional domains remains a fundamental challenge.
Existing option discovery methods are either confined to relatively simple domains, depend on handcrafted or quasi-symbolic representations, or offer little improvement over learning without options.
We present Wayfarer, a general, domain-agnostic, online deep RL agent that discovers options through Laplacian representation learning from high-dimensional observations and leverages them for control.
We show that the resulting options simultaneously improve exploration, accelerate credit assignment, and generalise effectively to unseen settings, enabling substantially faster learning of complex policies.
Wayfarer achieves state-of-the-art performance among single-stream agents on the most challenging Atari 2600 games, with the largest gains in games that require long-horizon exploration and strategic behaviour, such as Montezuma's Revenge and Private Eye.
Our code will be shared upon publication.}

\correspondence{\texttt{\{lintunen,machado\}@ualberta.ca}}

\newcommand{\pea}[1]{\textcolor[HTML]{4DA2D2}{#1}}
\newcommand{\peb}[1]{\textcolor[HTML]{B86D96}{#1}}
\newcommand{\pec}[1]{\textcolor[HTML]{008E68}{#1}}

\begin{document}

\maketitle

\section{Introduction}
\label{sec:introduction}

\noindent
\begin{wrapfigure}{r}{0.2175\textwidth}
  \centering
  \vspace{-\intextsep}
  \vspace{-2.7em}
  \includegraphics[width=\linewidth]{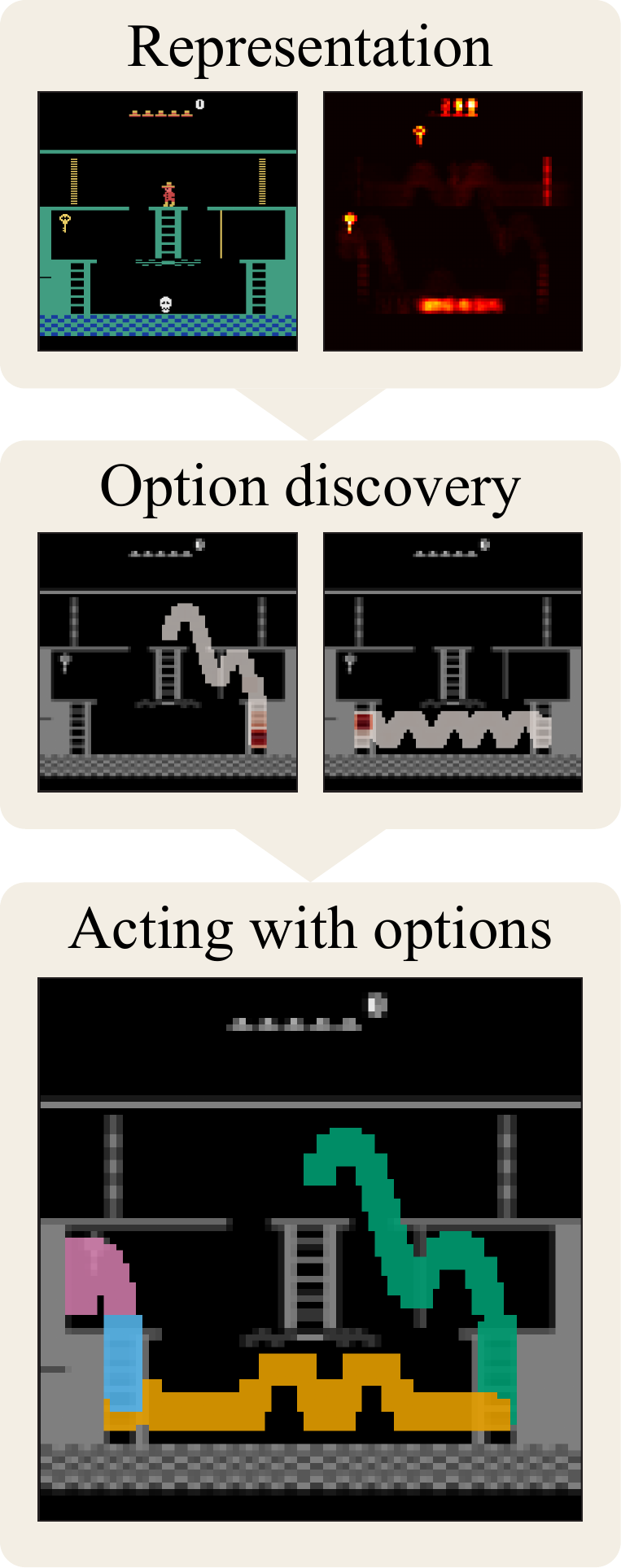}
  \captionsetup{
  format=plain,
  justification=raggedright,
  singlelinecheck=false
  }
  \vspace{-1.3em}
  \caption[Wayfarer learns a representation, derives options
  from it, and learns a policy over actions and options.]{%
  \textit{Wayfarer}\\
  learns a representation,\\
  derives options from it,\\
  and learns a policy over\\
  actions and options.%
  }
  \label{fig:frontpage}
  \vspace{-3em}
\end{wrapfigure}%
{\small
\noindent\textit{We must perceive in order to move,}
\textit{but we must also move in order to perceive.}\\[2pt]
\hspace*{\fill}--- J.\,J.~Gibson~\citeyearpar[p.~213]{gibson1979the}\par}
\medskip
Intelligent behaviour spans many time scales at once.
Imagine travelling to a conference: you book the trip, navigate an unfamiliar city, and, far down the hierarchy, coordinate innumerable muscle twitches.
Reinforcement learning~(RL) commonly formalises such decision-making in discrete time steps: at each step, the agent observes its environment, takes an action, and receives a reward.
Maximising this reward signal defines the agent's goal.
The time step is the fundamental unit of time in this formalism, yet many intelligent behaviours are built from what the formalism itself does not immediately supply: temporally extended courses of action, each chosen as a single decision, at multiple levels of temporal abstraction.

\emph{Options} \citep{sutton1999between,precup2000temporal} are central to such abstraction in RL.
An option is a skill, formalised as a policy paired with a termination condition, that the agent can use for planning, for credit assignment, or as an abstract action to follow---that is, acting according to the option's policy until it terminates.
An option abstracts away low-level actions and instead represents a temporally extended behaviour---crossing the room, descending the ladder---so decisions become fewer and the intervals between them longer.
Therein lies the long-recognised promise of options \citep[pp.~5--6]{klissarov2025discovering}: they facilitate exploration, because directed behaviour replaces step-by-step dithering \citep{machado2019efficient}; they accelerate temporal credit assignment, because outcomes are attributed to a few long decisions rather than many short ones \citep{sutton1999between}; and they enable generalisation, because a skill learned once can transfer to situations never seen before \citep{konidaris2007building}. %

In practice, however, options are often given, not discovered.
Deep RL agents now master real-time strategy games \citep{vinyals2019grandmaster}, navigate stratospheric balloons \citep{Bellemare2020autonomous}, and outplay elite athletes \citep{durr2026outplaying}, but where such systems act at multiple time scales, the abstraction is engineered by their designers rather than learned from experience.
Developing option-discovery methods that are effective at scale, in high-dimensional domains, remains a fundamental challenge.

Many existing discovery methods ground their options in the task reward \citep{vezhnevets2017feudal,bacon2017the} or in features given by the designer \citep{singh2004intrinsically,kulkarni2016hierarchical}.
But where non-zero rewards are sparse, methods grounded in reward have no signal to build on.
Handcrafted subgoals, providing their own signal, get around this.
Yet to give the agent its subgoals is to assume what should be discovered, and such a method is only as general as the subgoals it is given.
Finding truly general methods remains a major open question.

We tackle this question head-on by grounding option discovery in the structure of the environment, as captured by the \emph{Laplacian representation} \citep{mahadevan2005proto-value}.
The Laplacian representation maps observations to a set of features that captures how slowly they vary along the agent's experience.
Observations close in time map to nearby points in the representation space, and the smoothest features trace the environment's large-scale, slowest-evolving structure.
Laplacian-derived options \citep{machado2017a,machado2018eigenoption,jinnai2019discovering,jinnai2020exploration,wang2021towards,theophilou2025novel,mahadevan2007proto-value,farebrother2023proto-value,touati2023does,chandrasekar2026the,shehmar2026laplacian} thus need neither task rewards nor designer-given subgoals: they are grounded in the agent's observations alone. %

The Laplacian representation has been used to facilitate exploration \citep{machado2017a,machado2018eigenoption,jinnai2019discovering,jinnai2020exploration,wang2021towards,theophilou2025novel}, credit assignment \citep{mahadevan2007proto-value,farebrother2023proto-value}, skill transfer \citep{touati2023does,chandrasekar2026the}, and planning \citep{shehmar2026laplacian}; but each in isolation.
Moreover, the representation is typically learned offline, from a pre-collected dataset, which assumes that we already have the very coverage that only good exploration could have produced.
The exception that learns it online uses its options for exploration alone \citep{klissarov2023deep}.
Indeed, to our knowledge, no option-discovery method---whether grounded in reward, in given subgoals, or in the structure of the environment itself---has been shown to be able to effectively tackle exploration, credit assignment, and generalisation at once when processing high-dimensional observations, let alone doing so online, from a single stream of experience. %

We present \emph{Wayfarer}, a general, domain-agnostic, model-free agent that closes this gap by combining modern deep RL with Laplacian representation learning.
Wayfarer couples three networks trained simultaneously, online and off-policy, from a single stream of experience (\autoref{fig:frontpage}).
\textbf{(1)} A representation network approximates the Laplacian representation of an abstract, \emph{contingent-aware} feature space that represents what is under the agent's control \citep{watson1966the,bellemare2012investigating}.
\textbf{(2)} A skill network learns a repertoire of temporally extended behaviours.
\textbf{(3)} A main network learns to maximise reward over the action set augmented with these options.
The networks form a \textbf{virtuous loop}: the representation defines the options' intrinsic rewards; the options augmenting the agent's action set shape the experience it collects; and this experience is used when training the networks, so the representation and the options improve one another as the agent expands the frontier of what it knows. 

We show that Wayfarer delivers the promise of options: they improve exploration, accelerate credit assignment, and generalise to previously unseen settings, enabling substantially faster learning of complex policies.
Three phenomena underlie these gains.
\textbf{First}, the smoothest features of the representation come to identify regions in the agent's environment that are weakly connected, and learning options from this information enables the agent to navigate between them.
\textbf{Second}, these options allow the agent to quickly learn a rough policy sketch that, over time, is distilled and refined into a policy over low-level primitive actions.
\textbf{Third}, we show options discovered in a specific part of the environment are readily available and effective in other unseen parts of the environment.

Among agents trained on a single stream of experience, Wayfarer achieves the state of the art on the \emph{challenging set}: ten Atari 2600 games that demand hard exploration or long-horizon credit assignment, and on which progress has historically been slow \citep{badia2020agent57}.
It does so with a single hyperparameter configuration across games.
The largest gains appear where long-horizon exploration and strategic behaviour are required, as in \emph{Montezuma's Revenge} and \emph{Private Eye}.

\section{Reinforcement learning with Laplacian-based options}
\label{sec:background}

We consider an agent learning to achieve goals by interacting with its environment in a sequence of episodes.
The agent starts from an initial state, $s_0\in\gS$, receives an observation from the environment, $x\in\gX$, and takes an action from the set of available actions $a\in\gA$.
The environment responds by transitioning to an unobservable state, delivering to the agent a reward signal, $r\in\gR\subset\RR$, and a new observation $x'\in\gX$.
This interaction continues until the episode terminates, after which the agent starts again from an initial state.
The agent's goal is defined by the reward, and it learns a policy to maximise its return---the expected discounted sum of the rewards over time.
An action-value function predicts this return for an observation--action pair, which allows the agent to estimate how rewarding particular observation--action combinations are over the long run. %

In the high-dimensional setting, observations can contain substantial variation unrelated to the agent's actions: the agent can perceive information through its sensory stream that it does not control.
\citet{efroni2022provably} formalise this distinction: observations are emitted by latent states, each factoring into two components: an \emph{endogenous} component, whose transitions depend on the agent's actions, and an \emph{exogenous} (``noise'') component, which evolves independently.
We assume that such a dynamics factor exists, which motivates learning an \emph{agent-centric} representation \citep{islam2023principled}: a mapping from the agent's observations to their endogenous components---a \emph{contingent} feature space.

\emph{Options} extend the agent's decision making across time \citep{sutton1999between,precup2000temporal}.
Each option $o\in \gO$ has a policy over primitive actions and a termination function.
Under \emph{call-and-return} execution, selecting an option delegates control to that option's policy until termination; after termination, control returns to a higher-level policy over the options.
A primitive action is an option of length one, so $\gA\subset\gO$.

The transition dynamics of an environment give the state space a shape.
Under a fixed policy, they induce a weighted graph: states are nodes, and the edge weight $w_{s,s'}\geq 0$ between states $s$ and $s'$ reflects how directly the dynamics connect them.
The adjacency matrix $\mat{W}\in\RR^{\nS\times \nS}$, with $W_{s,s'}=w_{s,s'}$, contains these weights; the degree matrix $\mat{D}$ is the diagonal matrix of the row sums of $\mat{W}$; and the \emph{graph Laplacian} for symmetric weight matrices is given by $\la=\mat{D}-\mat{W}$.\footnote{We present the discrete Laplacian for ease of exposition, but use its abstract generalisation \citep{wu2019the,wang2021towards,gomez2024proper}.}
In the RL setting, the Laplacian can be defined as any matrix $\la=\id-f(\pp)$, where $\pp$ defines the probabilities of transitioning between states while following policy $\pi$, and where $f:\RR^{|\gS|\times|\gS|}\rightarrow \operatorname{Sym}_{|\gS|}(\RR)$ maps $\pp$ to a symmetric matrix \citep{gomez2024proper}.
The eigenvectors of $\la$ define the \emph{Laplacian representation}, capturing the principal directions of information diffusion on the graph.
The eigenvectors corresponding to the smallest eigenvalues trace the graph's large-scale temporal structure, varying slowly across strongly connected states \citep{mahadevan2005proto-value}.
The first eigenvector is constant; subsequent ones capture this structure ordered by smoothness (\autoref{fig:gridlaplacian}, \autoref{app:laplacian}).
With high-dimensional observations, the agent must approximate the Laplacian representation from experience.
One way to do this is to approximate it from observations using stochastic optimisation \citep{wu2019the,wang2021towards,gomez2024proper,ahmed2025online}.
We typically parameterise this function using a deep neural network; we refer to the outputs of such a network as \emph{eigenfunction} estimates.
We discuss specific objective functions in \autoref{sec:wayfarer}.

Differences in the representation across observations can be used for \emph{intrinsic rewards} \citep{oudeyer2007what,linke2020adapting,colas2022autotelic,aubret2023information,lidayan2025bamdp}, providing the agent with a learning signal independent of task reward.
Options trained to follow the gradient directions of the representation can discover routes between weakly connected regions \citep{machado2017a,machado2018eigenoption}, empowering the agent to reach states unlikely to be encountered by dithering with primitive actions. %

\emph{Deep action-value learning} provides a way to learn both these option policies and when to use them.
The Deep Q-Networks (DQN) algorithm \citep{mnih2013playing,mnih2015human-level} estimates action values from pixels, learning off-policy from experience replay \citep{lin1992self-improving}.
Rainbow \citep{hessel2018rainbow} extends DQN with double Q-learning \citep{vanhasselt2010double}, prioritised experience replay \citep{schaul2016prioritized}, a dueling Q-network \citep{wang2016dueling}, multi-step learning \citep{sutton1988learning}, distributional RL \citep{bellemare2017a}, and a noisy network \citep{fortunato2018noisy}.
Wayfarer uses these components with Implicit Quantile Networks (IQN) \citep{dabney2018implicit}.
IQN agents approximate distributional returns via learned quantile functions; such agents form an \emph{implicit} representation of the return distribution.

\section{Wayfarer}
\label{sec:wayfarer}

Wayfarer couples representation learning, option discovery, and task learning in a single online loop.
The agent begins by acting randomly and experiencing how the environment responds.
Henceforth, it instantiates the \emph{representation-driven option discovery} (ROD) cycle \citep{machado2023temporal,machado2025representation-driven}: the experience shapes a representation; the representation defines intrinsic rewards for option discovery; and the options shape what the agent experiences in the future. %
Early experience covers only a small part of a large state space, but even a rough representation supports options that reach states unlikely to be encountered by dithering with primitive actions.
We visualise the difference in \autoref{fig:pushfrontier} (\autoref{sec:experiments}).

The algorithm consists of three neural networks (\autoref{fig:wayfarerarchitecture}, \autoref{app:networks}).
The behaviour policy is determined by the \textbf{main network}, which learns from task rewards to estimate action values over both primitive actions and options.
When it selects an option, that option's policy controls the primitive actions until termination, with a fixed termination probability at each step.
The \textbf{skill network} learns a pair of options for each non-constant eigenfunction of the representation: one rewarded for increasing the value of the eigenfunction, the other for decreasing it.
The \textbf{representation network} learns these eigenfunctions, approximating the Laplacian representation of the agent-centric state---capturing slowly varying structure in the parts of the agent's observations that are under its control. %

All networks learn off-policy from a shared experience replay buffer, $\D$, and they all update within the same update step.
Our learning algorithm addresses three demands of this training loop: \textbf{(1)} extracting useful structure from high-dimensional observations; \textbf{(2)} transforming that structure into coherent, directed behaviours; and \textbf{(3)} learning when these behaviours help maximise task reward.

While interacting with the environment, the agent minimises four loss functions.
The main network learns action and option values to maximise task rewards (\autoref{sec:mainnetwork}), while the skill network (\autoref{sec:optionnetwork}) learns action values from intrinsic rewards induced by the representation learned by the representation network (\autoref{sec:repnetwork}).
Both the main and skill networks use a form of \emph{Q-learning} \citep{watkins1992q-learning}, combining rewards from replayed experience with bootstrapped estimates of future return.

For option discovery, we want the representation to capture temporal structure in aspects of the environment that the agent can influence.
To achieve this, the representation network regularises its encoder using \emph{multi-step inverse-dynamics prediction} \citep{gregor2016variational,efroni2022provably}, which discovers a feature space of information useful for inferring the agent's actions.
Using this encoder, a separate head in the network approximates the Laplacian representation through a variant of the \emph{spectral graph-drawing} objective \citep{koren2003on}, originally formulated as a constrained optimisation problem that minimises variation across connected states subject to orthonormality constraints.
This objective can be relaxed; the smoothness and constraint terms are estimated from experience, allowing the representation to be learned via stochastic optimisation \citep{wu2019the,wang2021towards,gomez2024proper,ahmed2025online}.
The two objectives train jointly, encouraging eigenfunctions that capture temporal structure from control-relevant information.\footnote{The main network has its own encoder and therefore the agent can still use information beyond its control---such as a countdown timer or passing traffic---in its higher-level decision making.}

\subsection{Acting over the option set}
\label{sec:mainnetwork}

To the agent, a primitive action is just another option in its option set, $a\in\gA\subset\gO$, %
so we denote both primitive actions and options by $o\in\gO$.
The main network estimates values over $\gO$, combining Rainbow's components with IQN.
Noisy layers in the dueling value and advantage streams contribute to exploration alongside the temporally extended behaviours induced by the options.

Let $Z(x,o)$ denote the return random variable associated with observation $x$ and option $o$, and let $F_{Z(x,o)}$ denote its cumulative distribution function.
We write $Z(x,o,\tau):=F_{Z(x,o)}^{-1}(\tau)$ for its quantile at level $\tau\in[0,1]$.
IQN approximates this quantile function with a \emph{universal value function approximator} \citep{schaul2015universal} that maps from $x$, $o$, and $\tau\sim U[0,1]$ to samples from the implicitly defined return distribution \citep[p.~4]{dabney2018implicit}.
A distortion function $g:[0,1]\rightarrow [0,1]$ transforms the sampled quantile levels, resulting in the \emph{distorted expectation} $Q_g(x,o) := \E_{\tau\sim U[0,1]}\big[ Z\left( x,o,g(\tau) \right) \big]$.
Different choices of $g$ give rise to a spectrum of \emph{risk-sensitive} policies.
Identity corresponds to risk-neutrality.

When no option is active, the agent selects the action or option with the highest estimate of the expected return, with noisy layers injecting exploration noise into the agent's estimates.
The agent's behaviour policy is risk-neutral.
To estimate the expectation, the agent draws $N_{\mathrm{sel}}$ independent quantile levels from $U[0,1]$:
\begin{align*}
\pi(x) := \argmax_{o\,\in\,\gO}\; \frac{1}{N_{\mathrm{sel}}} \sum_{i=1}^{N_{\mathrm{sel}}} \hinv\big( Z(x, o, \tau_i) \big).
\end{align*}
The value estimates are learned in a transformed space, using $\hsq(z):=\operatorname{sign}(z)\big(\sqrt{|z|+1}-1\big) + \varepsilon z$ for a small constant $\varepsilon$. Applying $h^{-1}$ before averaging returns them to the original scale.
This transformation is used as an alternative to reward clipping, reducing the scale of the optimisation target while retaining information about reward magnitude discarded by sign clipping \citep{pohlen2018observe,kapturowski2018recurrent,badia2020never,badia2020agent57}.

Selecting option $o$ delegates control to that option's policy, $\pi_o$, until it terminates.
The probability of termination is $P_{\mathrm{term}}$ in every non-terminal state, giving an expected duration of $\nicefrac{1}{P_{\mathrm{term}}}$ steps.

\paragraph{Learning from task rewards.}
The main network learns from prioritised experience replay using the truncated $n$-step return $G_t^n=\sum_{m=0}^{n-1}\gamma^{m}r_{t+m+1}$.
The online network selects the bootstrap action, and a target network \citep{mnih2015human-level} evaluates it.
With the online and target $Z$-functions parameterised by $\theta$ and $\theta^-$, respectively, the target is
\begin{align*}
y_j := \hsq\Big( G_t + \gamma^{n}\, \hinv\big( Z_{\theta^-}(x_{t+n}, o', \tau'_j) \big) \Big)\,,
\quad
o' := \argmax_{o\,\in\,\gO}\; \frac{1}{N_{\mathrm{sel}}} \sum_{i=1}^{N_{\mathrm{sel}}} \hinv\big( Z_\theta(x_{t+n}, o, \tau_i) \big)\,.
\end{align*}
The quantile-Huber loss \citep{huber1964robust} compares these targets with the online quantile estimates, and replay importance weights adjust each sample's contribution to correct for sampling bias (\autoref{app:hypers}).

In online ROD, options are derived from a representation that changes continually as the agent interacts with the environment.
Their policies therefore change too (the development visualised in \autoref{fig:pushfrontier}, \autoref{sec:experiments}), posing one of the biggest challenges in learning values over the discovered options.
In particular, a previously rewarding behaviour may not remain available through the same option.
Our aim is, therefore, to let expected task return guide which options to initiate and when, while distilling the useful behaviours into a policy over primitive actions as the options evolve.

To do such distillation, the agent learns as follows.
When the main network initiates an option, the replay sample stores that option.
At every other step, including subsequent steps during that option's execution, it stores the primitive action taken.
For simplicity, the main network updates both option and action values using the fixed $n$-step target above, bootstrapping regardless of whether the option has terminated.
Thus, task rewards train both option initiation and primitive action selection along the trajectories the options generate.
Empirically, the agent appears to favour options at the frontier of its experience and primitive actions as those regions become familiar (\autoref{fig:mreosr}, \autoref{app:eosr}).
Viewing $\D$ as a non-parametric model and replay as planning \citep{vanseijen2015a}, we can interpret this shift as the agent distilling, via planning, experience generated by options into a policy over primitive actions.

\subsection{From the representation to options}
\label{sec:optionnetwork}

While the main network learns about the expected return of such options, that is, the situations that are best for using options, the skill network learns the options' policies.
To do this, the network uses intrinsic rewards induced by the learned representation as goals that options learn to maximise.
In particular, these intrinsic reward functions are derived from the Laplacian representation, which the agent approximates using $d$ eigenfunctions. Each eigenfunction, $u_i$, after discarding the first, $i\in\{2,\cdots,d\}$, gives rise to two intrinsic reward functions:
\begin{align*}\label{eq:intrewards}
    r^{i+}(x_t,x_{t+n}):=u_i(x_{t+n})-u_i(x_t)\,,
    \qquad
    r^{i-}(x_t,x_{t+n}):=-\big(u_i(x_{t+n})-u_i(x_t)\big)\,.
\end{align*}
The multi-step horizon $n$ is shared with the main network.
The one-step intrinsic rewards telescope, meaning the intermediate terms cancel, so each reward is an undiscounted sum of one-step eigenfunction changes over this horizon.
The rewards are computed with the current representation, so they change as the agent learns more about the environment's structure.

The skill network learns $K=2(d-1)$ options, off-policy, from the agent's experience replay buffer.
A shared convolutional encoder thus feeds $K$ IQN heads, each estimating primitive-action values for one intrinsic reward, and learning from uniformly sampled experience replay using $n$-step updates.
Each option's policy acts greedily over $\gA$ with respect to the mean of its estimated value quantiles.
All $K$ heads update at every update step, and the shared encoder descends on their averaged gradient. %

\subsection{From observations to the representation}
\label{sec:repnetwork}

The Laplacian captures changes in slowly evolving aspects of the agent's experience \citep{sprekeler2011on}.
These changes often correspond to visiting states that are hard to reach---rare occurrences in the agent's data stream---and thus help guide exploration.
However, this can create a trap: a slowly varying eigenfunction need not capture variation contingent on the agent's actions.
A timer can count down or traffic can move slowly across the agent's field of view regardless of what the agent does.
An option may then receive reward for these changes without influencing them.
Wayfarer therefore learns an abstract agent state and a Laplacian representation of the graph induced by that state.

To do this, Wayfarer couples Laplacian representation learning with a prediction problem that requires the agent to learn about the effects of its actions on its future observations.
The representation network consists of a convolutional encoder parameterised by $\psi$ shared by a Laplacian head parameterised by $\phi$ and an inverse-dynamics prediction head by $\chi$.
The two heads output eigenfunction estimates, $\rvu\in \RR^d$, and probabilities over actions, $\hat{\rva}\in [0,1]^{|\gA|}$, respectively.
We write $u_i$ to denote the $i$-th output of the Laplacian head and $\hat{a}_i$ to denote the $i$-th output of the inverse-dynamics head.

\paragraph{Agent-centric regularisation.}
The agent predicts the first action of a multi-step transition from the concatenated ($\oplus$) state-feature vectors of two observations separated by $\Delta$ steps \citep{efroni2022provably}.
Specifically, the agent replays transitions $(x_t,a_t,x_{t+\Delta})$ from $\D$, with $P(\Delta)\propto\gamma_{\mathrm{rep}}^{\Delta-1}$ truncated at the episode boundary, to minimise the cross-entropy loss:
\begin{equation*}
\begin{aligned}
    \loss_{\mathrm{inv}}(\psi,\chi)
    &:= \E_{(x_t,a_t,x_{t+\Delta})\sim P_{\mathrm{rep}}}
    \big[-\log q_\chi(a_t\mid x_t,x_{t+\Delta})\big]\,,\\
    q_\chi(\cdot\mid x_t,x_{t+\Delta})
    &:= \softmax\Big(f_\chi\big(f_\psi(x_t)\oplus f_\psi(x_{t+\Delta})\big)\Big)\,,
\end{aligned}
\end{equation*}
where $P_{\mathrm{rep}}$ is the joint distribution induced by sampling $(x_t,a_t)$ uniformly from $\D$ and $\Delta\sim P(\Delta)$.

\paragraph{Laplacian representation learning.}
The Laplacian objective seeks diverse and smooth eigenfunctions. %
Wayfarer uses the \emph{Augmented Lagrangian Laplacian Objective} (ALLO) \citep{gomez2024proper}:
\begin{equation*}
    \max_{\dual}\;\min_{\phi}\;
    \sum_{i=1}^{d} \ip{u_i,Lu_i}
    + \sum_{j=1}^{d}\sum_{k=1}^{j}\beta_{jk}\big(\ip{u_j, \sg{u_k}} - \delta_{jk}\big)
    + b \sum_{j=1}^{d}\sum_{k=1}^{j}\big(\ip{u_j, \sg{u_k}} - \delta_{jk}\big)^2\,.
\end{equation*}
The first term is estimated from observation pairs $(x_t,x_{t+\Delta})$ replayed from $\D$, with $\Delta\sim P(\Delta)$. %
This term penalises changes across observations over time.
The remaining terms enforce orthonormality through Lagrange multipliers $\beta_{jk}$ and a quadratic penalty of weight $b>0$.
The Kronecker delta $\delta_{jk}$ specifies the target inner product: one when $j=k$ and zero otherwise.
Both constraint terms are asymmetric: the inner sum stops at $j$, and the stop-gradient $\sg{\cdot}$ treats $u_k$ as constant under differentiation.
This asymmetry makes the Laplacian eigenfunctions the only stable equilibrium point, avoiding convergence to arbitrary rotations of them \citep{gomez2024proper}.

Wayfarer holds $b$ fixed and applies a symmetric log transformation to the representation outputs.
Both choices address the changing data distribution when learning from online experience: novel observations can produce large outputs and constraint errors, which a growing penalty would amplify.
The output transformation compresses extreme values, thereby stabilising the gradients of the loss. %

Writing what is inside ALLO's max-min objective as $\loss_{\mathrm{lap}}$, the joint representation objective is
\begin{equation*}\label{eq:repjointobjective}
    \max_{\dual}\;\min_{\psi,\,\phi,\,\chi}\; \loss_{\mathrm{lap}}(\psi,\phi;\dual,b) \;+\; w_{\mathrm{aux}}\, \loss_{\mathrm{inv}}(\psi,\chi)\,,
\end{equation*}
where $w_{\mathrm{aux}}$ weights the auxiliary objective.
Only the shared encoder $\psi$ receives both gradients.
The resulting representation is encouraged to capture temporal structure while retaining information useful for inverse dynamics.
Our experiments quantify the effect of this regularisation on the agent's performance, and \autoref{fig:freewayattribution} (\autoref{app:representation}) visualises the difference side-by-side.

\paragraph{Closing the loop.}
All three networks update in the same cycle.
Each has its own encoder: experience and intrinsic rewards connect the networks.
The main network's choices shape experience; this experience refines the representation; and the representation is used for learning options that feed back into the main network's choices.
Following this virtuous loop, Wayfarer learns from its experience, online, while learning how to shape it, from a single stream of experience. %

\section{Experiments}
\label{sec:experiments}

We evaluate Wayfarer on the ten Atari 2600 games of the \emph{challenging set} \citep{badia2020agent57} in the Arcade Learning Environment (ALE) \citep{bellemare2013the} with sticky actions, the full action set, and no access to life information \citep{machado2018revisiting}.
We first compare Wayfarer's performance with state-of-the-art deep RL agents, Rainbow \citep{hessel2018rainbow}, IQN \citep{badia2020agent57}, and DreamerV3 \citep{hafner2025mastering}, then zoom in to examine what the discovered options contribute.

\subsection{Experimental protocol}
\label{sec:atari2600challenging}

\begin{figure}[htbp]
    \centering
    \includegraphics[width=0.095\textwidth, height=0.1247\textwidth]{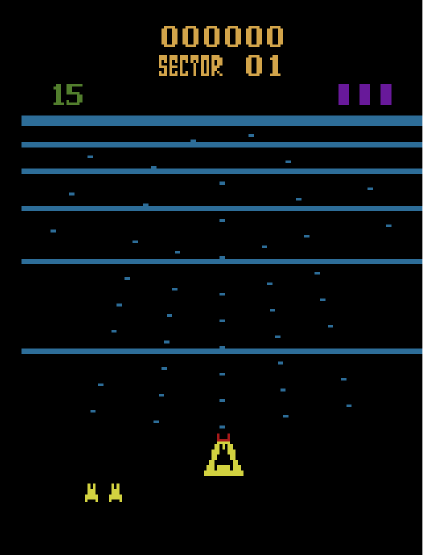}%
    \hfill
    \includegraphics[width=0.095\textwidth, height=0.1247\textwidth]{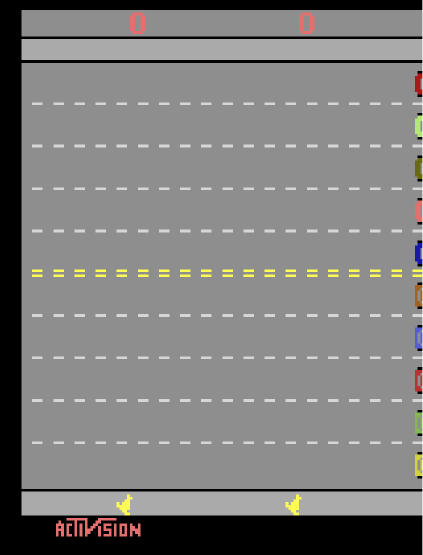}%
    \hfill
    \includegraphics[width=0.095\textwidth, height=0.1247\textwidth]{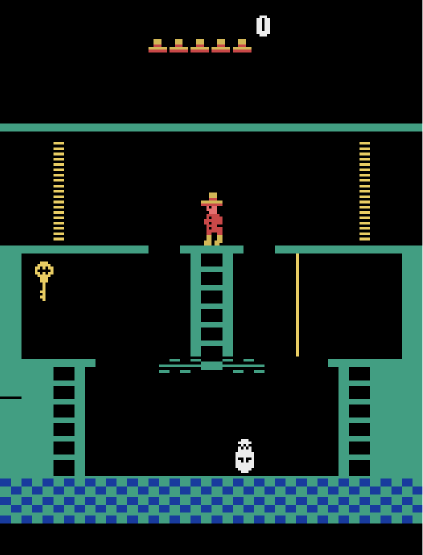}%
    \hfill
    \includegraphics[width=0.095\textwidth, height=0.1247\textwidth]{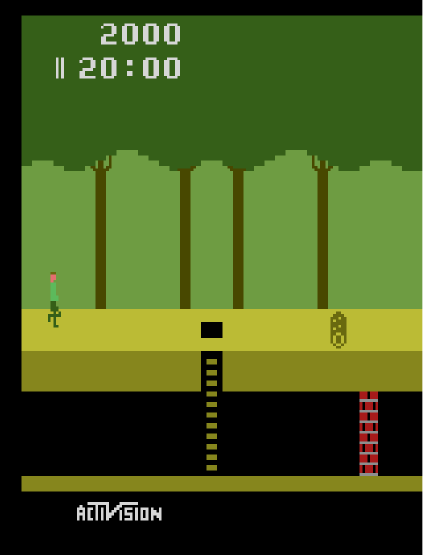}%
    \hfill
    \includegraphics[width=0.095\textwidth, height=0.1247\textwidth]{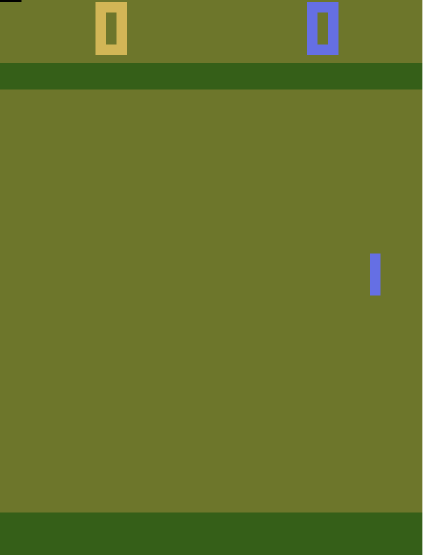}%
    \hfill
    \includegraphics[width=0.095\textwidth, height=0.1247\textwidth]{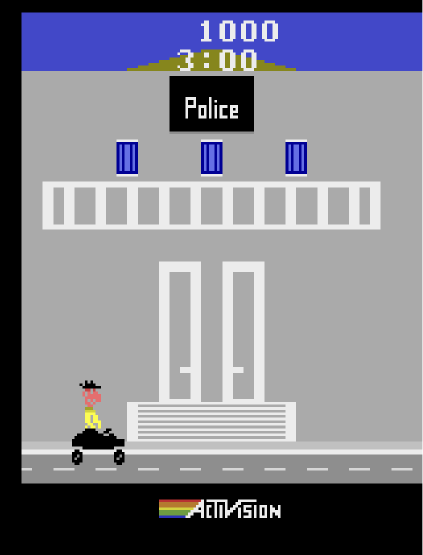}%
    \hfill
    \includegraphics[width=0.095\textwidth, height=0.1247\textwidth]{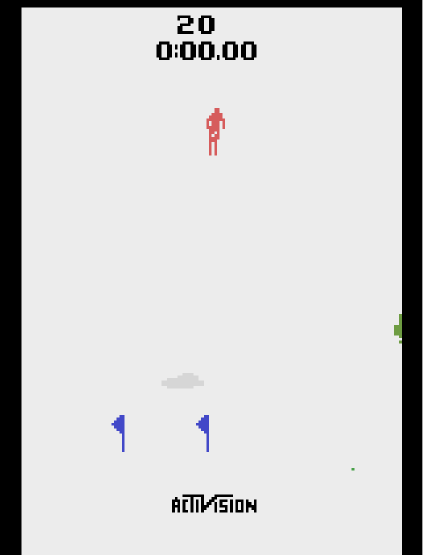}%
    \hfill
    \includegraphics[width=0.095\textwidth, height=0.1247\textwidth]{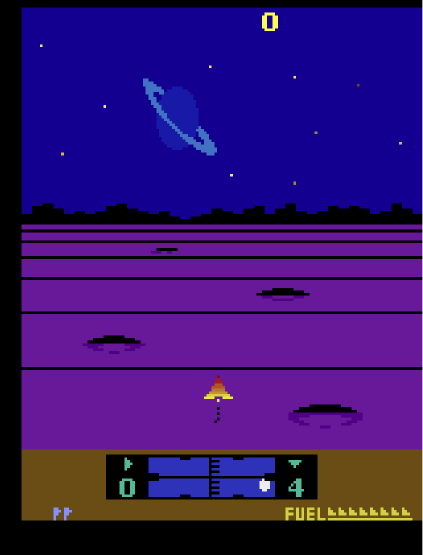}%
    \hfill
    \includegraphics[width=0.095\textwidth, height=0.1247\textwidth]{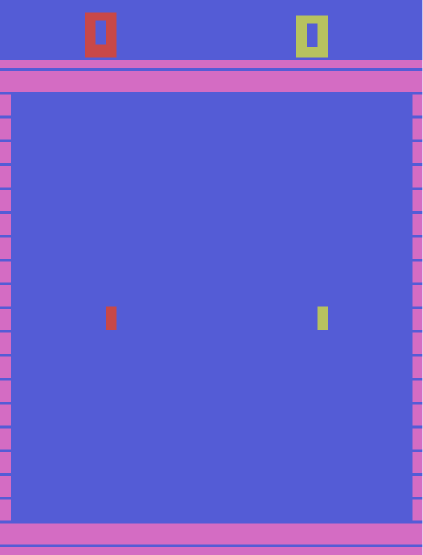}%
    \hfill
    \includegraphics[width=0.095\textwidth, height=0.1247\textwidth]{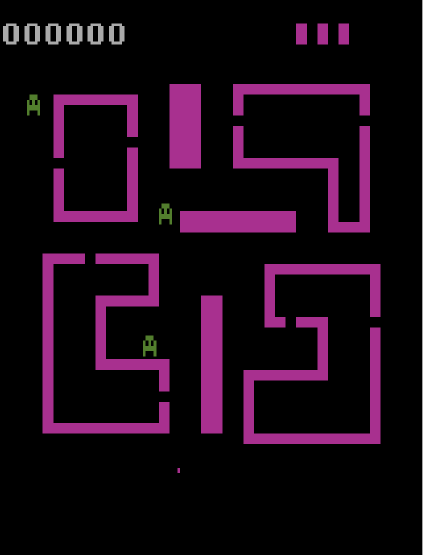}
    \caption{The Atari 2600 \emph{challenging set}.
    From left to right: \emph{Beam Rider}, \emph{Freeway}, \emph{Montezuma's Revenge}, \emph{Pitfall!}, \emph{Pong}, \emph{Private Eye}, \emph{Skiing}, \emph{Solaris}, \emph{Surround}, and \emph{Venture}.
    The set was assembled by \citet{badia2020agent57} to highlight games where progress has historically been much slower than average due to challenging exploration and credit assignment problems.
    Designed for entertainment, these games offer test beds of intelligent behaviour free of the experimenter's design bias \citep{bellemare2013the,allen2024using}.}
    \label{fig:chalsetframes}
\end{figure}

We train Wayfarer with a single hyperparameter configuration across games, using five runs per game for 200M frames in the main experiments and three for 100M frames in the ablations.
In aggregating results, each run contributes its mean episodic return over the final $10\%$ of training, without separate evaluation phases.
Wayfarer is a \emph{single-stream} agent: one learner, acting in one instance of the environment, learning online from the experience it generates.
Rainbow and IQN provide established, state-of-the-art single-stream, model-free baselines; Dreamer provides an additional state-of-the-art reference beyond the model-free class that learns from imagined experience and collects data from parallel environment instances, placing Wayfarer's performance in a broader context.
We report \emph{human-normalised scores} \citep{mnih2015human-level} using the median, interquartile mean (IQM), mean, and optimality gap, with 95\% bootstrapped confidence intervals \citep{agarwal2021deep}.
Further experimental details on hyperparameters, evaluation, and baselines are provided in Appendices \ref{app:hypers}, \ref{app:results}, \ref{app:baselines}, respectively.

\subsection{Results and ablations}

\begin{figure}[h]
  \centering
  \includegraphics[width=1.00\linewidth]{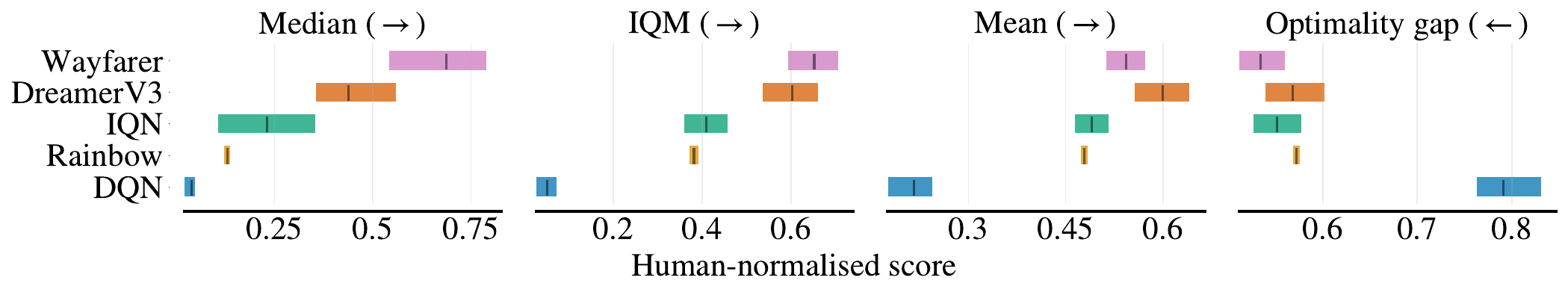}
  \caption{Aggregate metrics on the Atari 2600 challenging set, over 200M frames, with 95\% CIs.
  Higher mean, median, and IQM and lower optimality gap are better.
  We report results on ten games with five runs per game, except for DreamerV3, for which we use four---the minimum available number of published runs available across games.\looseness=-1}
  \label{fig:challengingresults}
\end{figure}

\autoref{fig:challengingresults} shows that \textbf{(1)} Wayfarer outperforms the model-free baselines; \textbf{(2)} it reaches the performance of Dreamer; and \textbf{(3)} all agents struggle on this benchmark, a testament to its difficulty.
These results show that there is significant room for single-stream agents to improve on this set of games. %

The per-game results help explain these differences.
Wayfarer's performance improves with the robustness of the summary statistic (mean 0.54, IQM 0.65, median 0.69), suggesting that its performance is pulled down by outliers where it performs poorly and that it performs robustly otherwise.
Of the five games where the human performs at least as well as the agents, Wayfarer outperforms its fellow agents in Montezuma's Revenge and Private Eye, Dreamer in Solaris, and Rainbow in Skiing; no agent achieves a positive return in Pitfall!.
Dreamer scores exceptionally well in Beam Rider, contributing to its advantage under the mean.
Wayfarer leads in Venture.
We provide per-game training curves in \autoref{fig:pergame} and reference scores in \autoref{table:challengingresultsiqm} (\autoref{app:results}). %

\paragraph{What are options good for?}
Recall the long-recognised promise of options that they improve exploration, because coherent, temporally extended behaviours replace step-by-step dithering; accelerate credit assignment, because credit can be assigned to whole behaviours; and generalise to unseen settings, allowing skills learned in one setting to be reused in others.
While prior work  has often focused on demonstrating these benefits in isolation, Wayfarer reaps all three at once, enabling substantially faster learning.
Next, we provide anecdotal evidence to support our claims. %

\paragraph{Options improving exploration.}
Most RL algorithms rely on local, single-step exploration approaches.
Wayfarer mixes temporally extended exploration, designed to take the agent to the states that seem far away from the agent, with local exploration designed to provide coverage of the states around the agent's location.
To demonstrate how options discovered by Wayfarer improve exploration, we disabled option termination to inspect the temporally extended behaviours they had learned.
We draw examples from two games among the hardest to explore: Pitfall! and Private Eye.

\begin{wrapfigure}{r}{0.45\textwidth}
  \centering
  \vspace{-\intextsep}
  \vspace{-0.1em}
  \includegraphics[width=\linewidth]{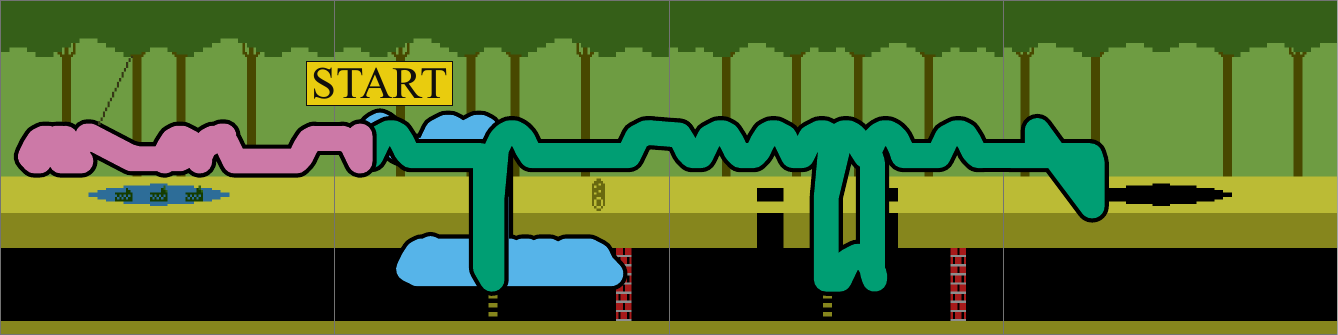}
  \captionsetup{font=small}
  \vspace{-2em}
  \caption{Three options discovered by Wayfarer in Pitfall!.
  One option (\peb{pink}) travels left, crossing a deadly puddle in a coordinated sequence.
  Another (\pec{green}) travels right, crossing holes and using ladders where needed. The third (\pea{blue}) underground.}
  \label{fig:pitfallexplore}
  \vspace{-2em}
\end{wrapfigure}
First, \autoref{fig:pitfallexplore} shows three options in Pitfall!.
To our knowledge, no single-stream agent, including Wayfarer, has achieved a positive return in the game.
Yet, by $20\%$ of the training budget, the agent has discovered skills that master non-trivial elements of the game.\customfootnote{The triangle symbol indicates that a corresponding video is available in the accompanying YouTube playlist:\\ \url{https://www.youtube.com/playlist?list=PLK1uRpOil4Yo}.}
One option crosses a deadly puddle; another leaps over holes in the ground, climbing back up each time it falls underground.
The third travels underground, a direction not covered by the other two. \looseness=-1

\begin{figure}[tpb]
    \centering
    \includegraphics[width=1.00\linewidth]{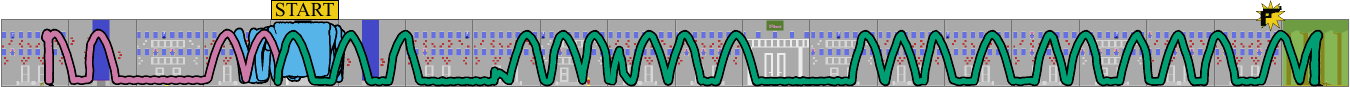}
    \caption{A single option discovered by Wayfarer, in Private Eye, at successive stages of training. The option corresponds to the smoothest eigenfunction, which captures the slowest-changing structure.
    The option is initialised as random (\pea{blue}), quickly expands coverage (\peb{pink}), and soon reaches a hard exploration-milestone (\pec{green}): the option obtains a key item in the investigator's case (gun).} %
    \label{fig:pushfrontier}
\end{figure}

Second, in Private Eye, a \gapuline{single} option learns to navigate streets, parks, and secret passages while avoiding dead ends for almost two minutes of gameplay.$^\triangle$
\autoref{fig:pushfrontier} shows this option from three earlier points in training.
By $6\%$ of the training budget, the option has reached a hard exploration-milestone: it directs the investigator all the way to the gun, a key milestone in solving the task.
In comparison, obtaining the gun takes $\sim$600 correctly sequenced decisions using primitive actions.

\paragraph{Options accelerating credit assignment.}
To evidence that options accelerate credit assignment, we compare Wayfarer's performance against a variant of it that does not learn option values, using the discovered options only as part of the agent's exploratory step.
This variant is based on Deep Covering Eigenoptions (DCEO) \citep{klissarov2023deep}.
DCEO agents use options as part of the agent's exploratory step while following an $\eps$-greedy policy.
When exploring, an option is chosen at random and its policy followed until termination.
To match DCEO, we change Wayfarer's configuration to use an $\epsilon$-greedy policy and the main network to only learn action values over primitive actions.
The results (\emph{Wayfarer 100M} versus \emph{no option values} in \autoref{fig:challengingablations}) clearly favour Wayfarer.

These results may be explained in part by the pattern in \autoref{fig:mreosr} (\autoref{app:eosr}), which visualises empirical option-selection rates over training checkpoints and rooms from a run of Montezuma's Revenge.
The heatmaps show that the agent has a strong preference for options shortly after encountering a new room: the pattern suggests that the agent is using the options to expand its frontier and then distils a fine-grained policy over primitive actions from the data generated by the options.

The bottom panel of our front-page figure (\autoref{fig:frontpage}) further exemplifies how options support credit assignment.
The panel shows how options discovered in the first room of Montezuma's Revenge can be stitched together to reach the key via a handful of high-level decisions, compared to $\sim$70 decisions via primitive actions.$^\triangle$
Even with a fixed update horizon, these directed sequences of actions effectively shorten the delay of the agent's reinforcement, likely helping the agent assign credit. %

\begin{figure}[htpb]
    \centering
    \includegraphics[width=1.00\linewidth]{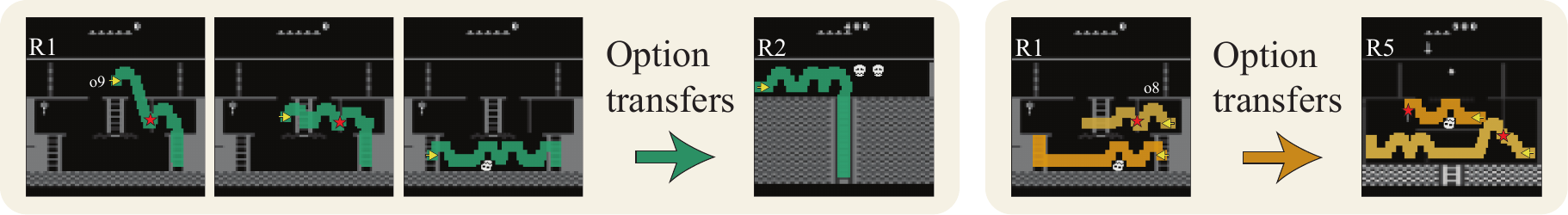}
    \caption{Options discovered in the first room of Montezuma's Revenge use ropes and avoid deadly skulls in unseen rooms without further learning.
    \figleft{} Option \#9 moves right and descends ladders.
    \figright{} Option \#8 moves left, also using a rope-like pole along the way.
    Starting points/directions are marked with yellow arrows, ropes and poles with red stars, and deadly enemies with skull icons.}
    \label{fig:optionsgeneralise}
\end{figure}

\begin{figure}[tpb]
  \centering
  \includegraphics[width=1.00\linewidth]{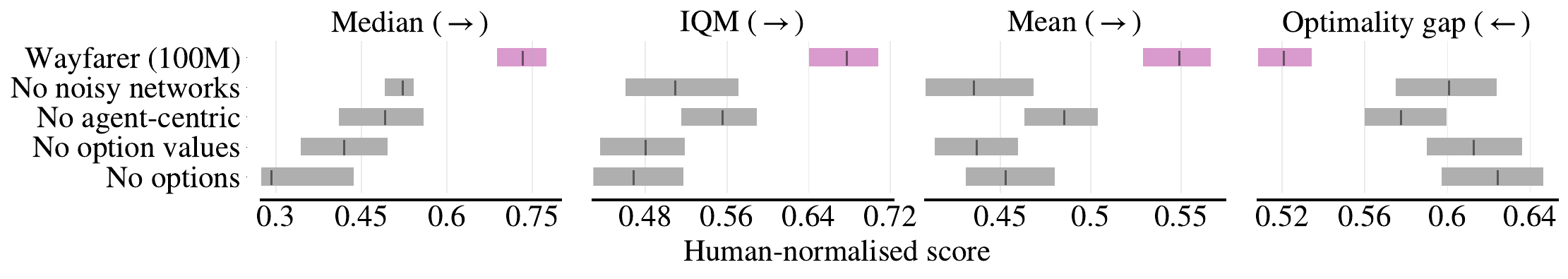}
  \caption{Aggregate metrics on the Atari 2600 challenging set, over 100M frames, with 95\% CIs. Higher mean, median, and IQM and lower optimality gap are better. Ten games; three runs per game.} %
  \label{fig:challengingablations}
\end{figure}

\paragraph{Options generalising to unseen settings.}
To show that options generalise to unseen settings, we study how the discovered options behave in regions of Montezuma's Revenge not yet experienced by the agent.
We choose a checkpoint at $0.4\%$ of the training budget, when we can be certain that the agent has experienced only the first room, and show how two selected options from this checkpoint transfer to unseen parts of the game (\autoref{fig:optionsgeneralise}).

Option \#9 moves Panama Joe to the right and down the ladder, a behaviour consistent from multiple starting states.
The skill remains the same in the next room. %
Option \#8 moves Panama Joe to the left via any ropes that are encountered along the way.
The skill remains the same in room five: from multiple starting states, the option travels left and is attracted to ropes and a rope-like pole.
Both options also allow Panama Joe to avoid deadly enemies (skulls) in unseen rooms, a valuable skill in a game where dying is easy.
A map of the game provides spatial context (\autoref{fig:mrlevel1}, Appendix~\ref{app:montezuma}).

\paragraph{Ablations.}
Our results suggest that options discovered by Wayfarer deliver on their promise: they improve exploration, accelerate credit assignment, and transfer to unseen settings.
Next, we examine how they interact with Wayfarer's design choices.
Specifically, we ask four questions.
\textbf{(1)} Can our good performance relative to baselines be attributed to options, or is it mostly explained by other parts of our algorithm?
\textbf{(2)} How important is learning the Laplacian from the agent-centric state as opposed to from raw observations?
\textbf{(3)} Does learning option values actually help or hinder performance compared with using the options only for exploration with an $\epsilon$-greedy policy?
\textbf{(4)} How does exploration through noisy value estimates compare with $\epsilon$-greedy exploration when option values are learned in both cases?
We tuned our ablated configurations; this is discussed in \autoref{app:baselines}. %

We make the following observations from the aggregate results in \autoref{fig:challengingablations} and the per-game training curves in \autoref{fig:pergameabla} (\autoref{app:results}).
\textbf{(1)} Learning options contributes substantially to Wayfarer's success: at 100M frames, a 2.51x improvement according to the median, 1.44x according to the IQM, and 1.21x according to the mean.
\textbf{(2)} Learning the Laplacian representation of an abstract, contingent-aware feature space is helpful.
We illustrate the difference using feature attribution in \autoref{app:representation}, which makes this result rather intuitive.
\textbf{(3)} Using options only for exploration performs competitively in the games that are hardest to explore but performs worse overall.
\textbf{(4)} The noisy network helps the agent compared with a prescribed $\epsilon$-greedy exploration schedule.
One possible explanation is that under $\epsilon$-greedy the agent may be forced to commit to temporally extended behaviour even when it is not needed.
Note that, with $\epsilon$-greedy exploration in both configurations (\emph{no noisy networks} and \emph{no option values}), learning option values plausibly improves aggregate performance slightly, which is consistent with our interpretation that options accelerate credit assignment. %

\section{Conclusion}
\label{sec:conclusion}

We have introduced Wayfarer, an online deep RL agent that discovers and learns to use options from high-dimensional observations.
On the Atari 2600 challenging set, the options discovered by Wayfarer deliver on their promise: they improve exploration, accelerate credit assignment, and generalise to unseen settings.
With these options, Wayfarer learns faster and achieves higher aggregate returns, surpassing the single-stream baselines IQN and Rainbow and matching the parallelised, model-based DreamerV3.
These behaviours emerge from a single stream of interaction with the environment, without domain-specific knowledge or semantic supervision.

Making Wayfarer work required tackling problems such as learning option values with non-linear function approximation \citep{kotamreddy2025a}, stabilising online Laplacian representation learning \citep{ahmed2025online}, and fusing Laplacian representation learning with agent-centric representation learning \citep{efroni2022provably,lamb2023guaranteed,islam2023principled}.
Throughout the paper, we have discussed the work most closely related to ours; since the option-discovery literature is vast, we include an extended related work in \autoref{app:extendedrelated}.
Based on our experiments, incorporating memory beyond frame stacking, sustaining plasticity (\autoref{app:nap}), and planning with discovered options \citep{sutton2023reward-respecting} offer clear directions for extending this work.
We conclude that Wayfarer is the beginning of something bigger: a prototype for a general agent that reasons at multiple levels of temporal abstraction and continually discovers new knowledge from experience in the environment.

\newpage

\subsection*{Acknowledgments}

We thank Diego Gomez for helping us understand the subtleties of ALLO; Nadia Ady, Jesse Farebrother, and Martin Klissarov for reading and commenting on parts of the manuscript; and Prabhat Nagarajan for discussions about learning Laplacian-based options when aspects of the environment's dynamics are beyond the agent's control.

Computational resources were provided by the Digital Research Alliance of Canada and the Aalto Science-IT project.

This research was supported in part by the Natural Sciences and Engineering Research Council of Canada (NSERC) and the Canada CIFAR AI Chairs Program.

Erik was supported in part by the Research Council of Finland's NEXT-IM project, by NSERC through a Canada Impact+ Research Training Award, and by the University of Alberta's Faculty of Graduate \& Postdoctoral Studies through the President's Doctoral Prize of Distinction.
Erik also received support through a Researchers Abroad grant from the KAUTE Foundation, the Walter Ahlstr\"om Foundation, the Foundation for Economic Education, and the Nokia Foundation.

\subsection*{AI use statement}

In this work, we used generative AI tools to polish our writing and to support most tasks involving the production of code (implementation, figures).
We take responsibility for the final content of this work, including text, claims or artifacts produced with the aid of generative AI.

\subsection*{Reproducibility statement}

We have made our best effort to ensure our work is reproducible.
\autoref{app:networks} illustrates our network architecture, \autoref{sec:wayfarer} describes our learning algorithm, and \autoref{app:hypers} lists our hyperparameters.
\autoref{app:results} details our score-aggregation procedure and the random and human reference scores used to compute human-normalised scores.
We will publish our code upon publication of this paper.

\bibliographystyle{assets/unsrtnat}
\bibliography{references}

\newpage
\appendix

\section{Examples of the Laplacian representation}
\label{app:laplacian}

\begin{figure}[htbp]
    \centering
    
    \begin{subfigure}[b]{0.5\textwidth}
        \centering
        \includegraphics[width=\textwidth]{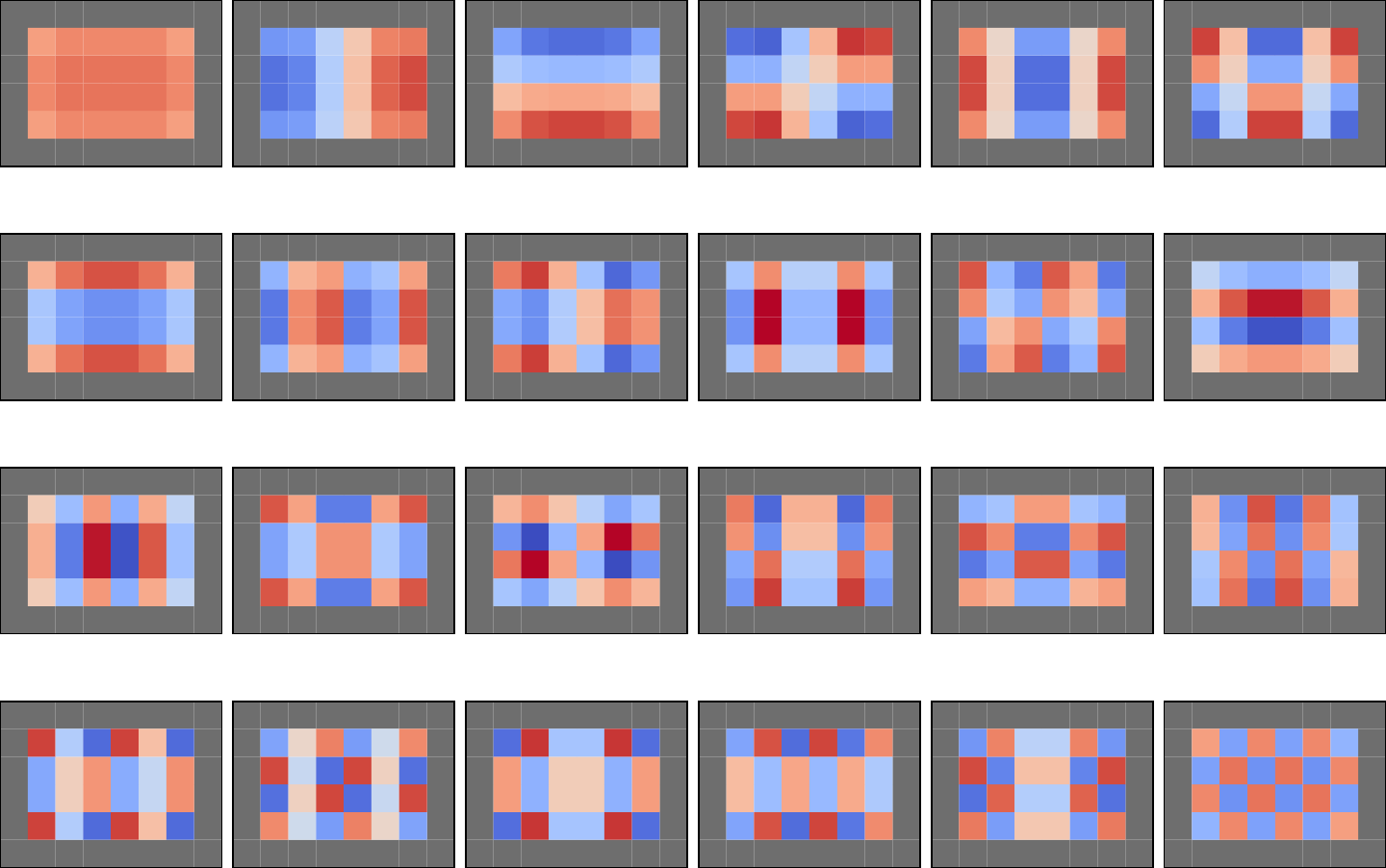}
        \caption{The 24 eigenvectors in increasing order of eigenvalue. The eigenvectors corresponding to the smallest non-zero eigenvalues capture large-scale variation across the grid.}
    \end{subfigure}
    
    \vspace{1.0em} %
    
    \begin{subfigure}[b]{0.5\textwidth}
        \centering
        \includegraphics[trim={0.6cm 0.1cm 0.2cm 0.6cm}, clip, height=2.5cm, keepaspectratio]{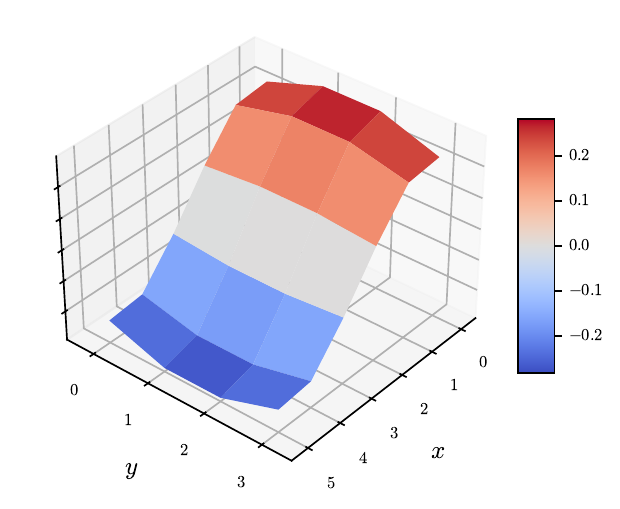}
        \caption{The second eigenvector plotted in three dimensions to better illustrate its topology.}
    \end{subfigure}
    
    \caption{Examples of Laplacian eigenvectors from a $4\times 6$ grid environment.}
    \label{fig:gridlaplacian}
\end{figure}

\section{The three Wayfarer networks}
\label{app:networks}

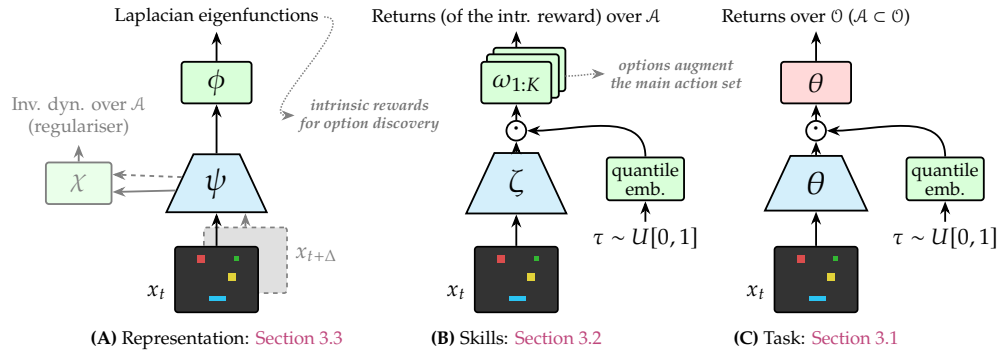
\begin{figure}[htpb]
    \centering
    \begin{adjustbox}{trim=11pt 3pt 4pt 1pt, clip, width=0.84\linewidth}
        \input{wf_networks.tex}
    \end{adjustbox}
    \caption{The three Wayfarer networks: \textbf{(A)} The agent-centric Laplacian representation network, whose outputs are used to provide intrinsic rewards for training \textbf{(B)} the skill network, which learns the options that augment the action set of \textbf{(C)} the main network learning the extrinsic task. Blue blocks are convolutional components; green ones are linear blocks; red ones are their noisy variants.}
    \label{fig:wayfarerarchitecture}
\end{figure}

Although the three networks in \autoref{fig:wayfarerarchitecture} are disjoint, they function synergistically throughout the representation-driven option discovery (ROD) cycle: the agent-centric Laplacian representation defines the intrinsic reward functions for option discovery, and the discovered options shape the agent's experiences.
These experiences are used to refine the representation, and the cycle continues in a virtuous loop.

\section{Investigating the agent's preferences}
\label{app:eosr}

To inspect the agent's decision making and value preferences, we compute the empirical option-selection rates in Montezuma's Revenge from experience and network checkpoints saved during training. First, we explain the basic mechanics of the game.

\subsection{Montezuma's Revenge: game mechanics and level one map}
\label{app:montezuma}

\begin{figure}[h]
  \centering
  \includegraphics[width=0.92\linewidth]{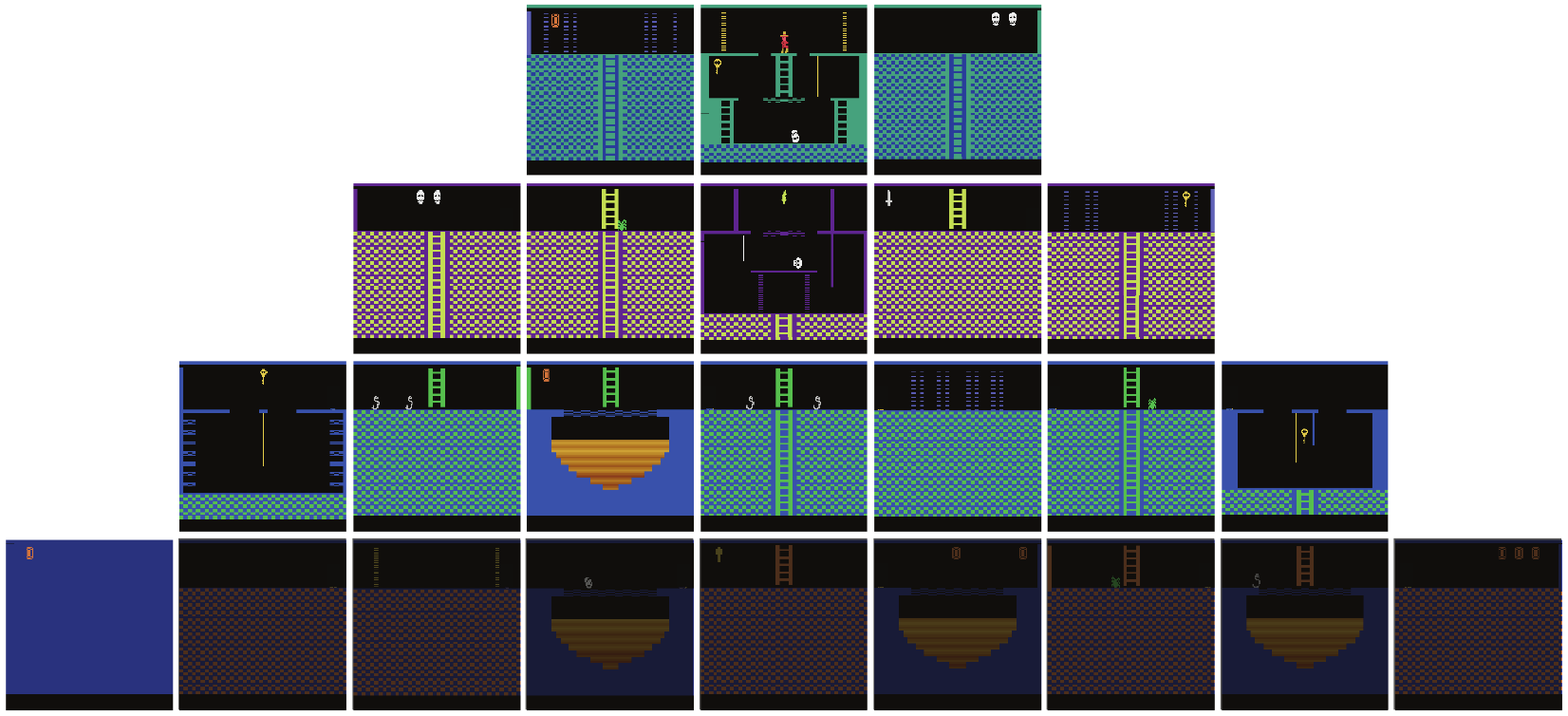}
  \caption{The first level of Montezuma's Revenge. Rooms that appear dark to the agent without a torch in its inventory are marked here with a black overlay. In game, the rooms are numbered from 0 to 23, row-major. The agent starts from room 1, and the level ends in room 15. To pass the level, at least three of the four keys have to be collected, which requires visiting at least 15 of the 24 rooms.}
  \label{fig:mrlevel1}
\end{figure}

\autoref{fig:mrlevel1} shows the first level of the game: the agent starts in the upper centre of the triangle map.
To proceed in the game, the agent must collect keys to open doors while avoiding various death traps.
Additionally, to accumulate more score, the agent can collect gems and slay enemies with a sword.
To see in the dark, the agent can seek out a hidden torch; and so on.
The agent can reach level two by making its way to the blue room in the bottom left of the triangle.
To pass the level, the agent has to collect three out of the four keys and refrain from using two of them until the third-to-last room.
To do this, the agent has to visit at least 15 out of the 24 rooms.

\subsection{Empirical option-selection rates}

We compute a quantity that we call the empirical option-selection rate (EOSR) as follows.
For each saved checkpoint, consisting of observations in the experience replay buffer and network parameters, the observations in the agent's experience replay buffer are grouped by room.
Each observation is passed through the main network, and EOSR is computed for each room as the fraction of observations for which the network selects an option over a primitive action.
If this quantity is higher than $\nicefrac{K}{A+K}$, where $K$ is the number of options and $A$ the number of primitive actions, then the agent has a preference for options over primitive actions relative to uniform selection over the augmented action set.
Conversely, if the quantity is lower, then the agent had a preference for primitive actions over temporally extended ones relative to the same baseline.

\begin{figure}[htbp]
    \centering
    \begin{subfigure}{0.49\textwidth}
        \centering
        \includegraphics[width=\linewidth, height=3.6cm, keepaspectratio]{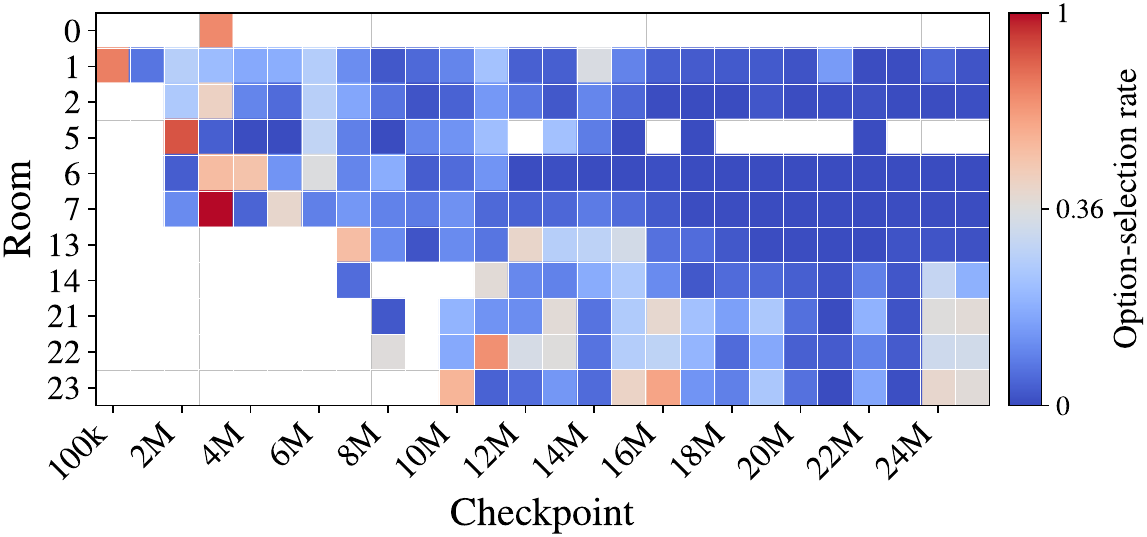}
    \end{subfigure}
    \hfill
    \begin{subfigure}{0.49\textwidth}
        \centering
        \includegraphics[width=\linewidth, height=3.6cm, keepaspectratio]{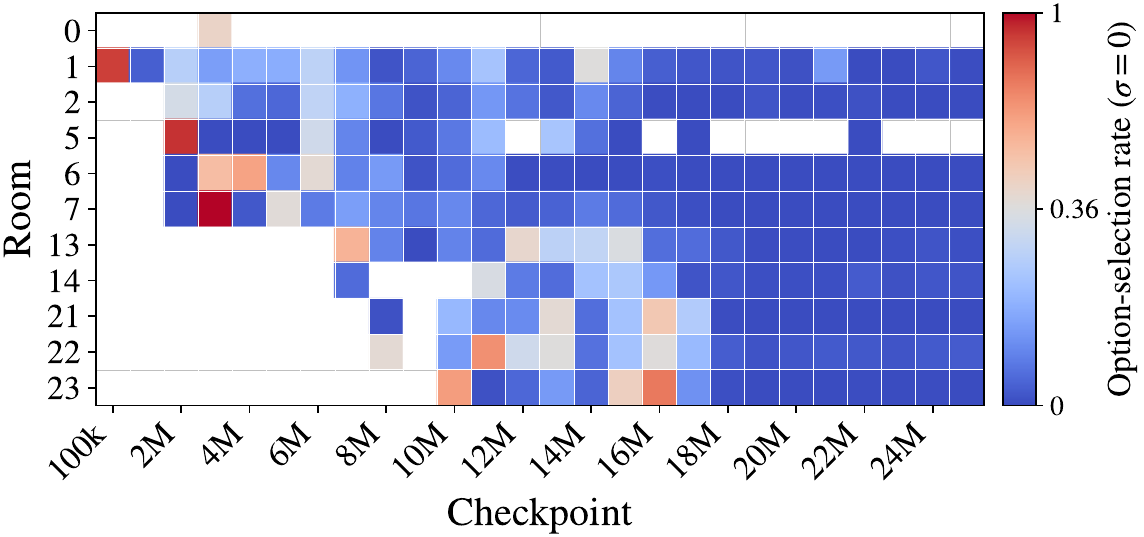}
    \end{subfigure}
    
    \caption{Empirical option-selection rates (EOSR) over the first 100M frames in Montezuma's Revenge: at each checkpoint, the observations in the agent's experience replay buffer are grouped by room; each observation is passed through the main network, and EOSR is the fraction of cases in which the network selects an option over a primitive action.
    Each cell quantifies whether the agent prefers options (\textcolor{red}{red}) or primitive actions (\textcolor{blue}{blue}) relative to uniform selection over the 18 primitive actions and 10 options ($\nicefrac{10}{28}\approx 0.36$), in a particular room of Montezuma's Revenge at a particular point in time during training.
    White cells mark rooms absent from that checkpoint's replay buffer.
    \figleft{} EOSR under the main network's noisy value estimates, averaged over 100 noise resamples per observation.
    \figright{} EOSR with noise disabled, using the $\argmax$ of the learned value estimates over temporally extended and primitive actions.}
    \label{fig:mreosr}
    \vspace{-0.9em}
\end{figure}

By dividing the observations of the experience replay buffer into rooms, and by doing so for each checkpoint, we can quantify whether the agent prefers options or primitive actions in a particular room of Montezuma's Revenge at a particular point in time during training.
\autoref{fig:mreosr} visualises the EOSR for both the main network's value estimates under noise (left), which is what defines the behaviour policy in training, and the same value estimates with noise disabled (right), thus visualising its learned preferences without NoisyNet noise.

\paragraph{Interpretation.}
Most RL algorithms rely on local, single-step exploration approaches.
Wayfarer mixes temporally extended exploration (options), designed to take the agent to the states that seem far away from the agent, with local exploration (noisy networks) designed to provide coverage of the states around the agent's location.
The two heatmaps suggest that noisy networks can often flip an action selection from primitive actions to options (and vice versa), supporting the agent's exploration.
Once a new room has been discovered, the agent develops a value preference for options over primitive actions.
These options then support the agent in distilling a fine-grained policy over its primitive actions to consistently reach that milestone.
Not only is this consistent with options accelerating credit assignment, but it also highlights a key advantage of learning option values: the agent can choose to use them if they help and ignore them if they do not.

\section{Wayfarer's hyperparameters}
\label{app:hypers}

\vspace{-0.9em}
\begin{table}[h]
\centering
\caption{Wayfarer's hyperparameters in the \emph{Arcade Learning Environment} \citep{bellemare2013the}.} %
\vspace{-0.1em}
\small
\begin{tabular}{@{}lll@{}}
\hline
\textbf{Symbol} & \textbf{Value} & \textbf{Meaning}\\
\hline
$d,\;K$ & $6,\;2(d-1)=10$ & number of eigenfunctions, eigenoptions\\
$\gamma,\;n$ & $0.99,\;3$ & discount factor, multi-step returns $n$\\
$N,\;N',\;N_{\mathrm{sel}}$ & $64,\;64,\;32$ & IQN online, target, selection quantiles\\
$m_{\tau},\;\kappa$ & $64,\;1$ & cosine embedding dims, Huber threshold\\
$B,\;B_{\mathrm{rep}}$ & $32,\;512$ & main \& skill minibatch, rep \& aux minibatch\\
& Adam \citep{kingma2015adam} & optimiser\\
$\eta_{\theta}=\eta_{\omega},\;\epsilon_{\mathrm{Adam}}$ & $5\!\times\!10^{-5},\;3.125\!\times\!10^{-4}$ & main \& skill IQN step size, Adam $\epsilon$\\
$\eta_{\phi}=\eta_{\chi}=\eta_{\dual}$ & $3\!\times\!10^{-4}\!\to\!3\!\times\!10^{-5}$ & Laplacian encoder, inverse-dynamics \& dual step size\\
$\dual_0,\;b$ & $-2.0,\;0.5$ & ALLO dual initialisation, barrier coefficient (fixed)\\
clip & $30$ & global-norm gradient clip (all optimisers)\\
$\sigma_{0}$ & $0.1$ & NoisyNet initial noise scale\\
$\alpha,\;\beta_{\mathrm{PER}},\;\epsilon_{\mathrm{PER}}$ & $0.2,\;0.4\!\to\!1.0,\;10^{-10}$ & PER exponent, IS exponent over training, priority offset\\
& proportional & PER variant\\
$P_{\mathrm{term}}$ & $0.2$ & option termination probability\\
$w_{\mathrm{aux}}$ & $1.0$ & inverse-dynamics auxiliary weight\\
$\gamma_{\mathrm{rep}}$ & $0.9$ & geometric discounting in $P(\Delta)$\\
update period & $4$ env steps & one gradient step per network\\
target period & $10{,}000$ env steps & $=2{,}500$ gradient steps\\
total steps & $5\!\times\!10^{7}$ env steps & $=200$M frames per training run\\
max ep. steps & $27{,}000$ env steps & $=108{,}000$ frames at frameskip $4$ ($\approx\;$30 min)\\
$|\D|,\;D_{\min}$ & $10^{6},\;20{,}000$ & replay buffer capacity, min. replay samples before training\\
\hline
\end{tabular}
\label{tab:hp}
\end{table}

\section{Results}
\label{app:results}

We measure mean episodic returns during training, without separate evaluation phases.
Each episode has a time limit corresponding to 30 minutes of gameplay.
For the main experiments, we report five runs per game for 200M frames; for the ablations, we report three runs per game for 100M frames.
Each run contributes its mean episodic return over the final $10\%$ of training to the aggregate results.

\paragraph{Score aggregation.}
We report \emph{human-normalised scores} \citep{mnih2015human-level}: zero corresponds to the performance of a random agent and one to the human reference reported by \citet{schrittwieser2020mastering}.
Following \citet{agarwal2021deep}, we summarise performance using the median, interquartile mean (IQM), mean, and optimality gap, with 95\% bootstrapped confidence intervals produced using \emph{rliable} \citep{agarwal2021deep}.

For the main comparisons, we report results on ten games with five runs per game, except for DreamerV3, for which we use four---the minimum number of published runs available across games.
Using the same number of runs per game within each method keeps games equally represented when pooling scores.
For DreamerV3, we took the first four seeds in the dataset.

\begin{figure}[h]
  \centering
  \includegraphics[width=1.00\linewidth]{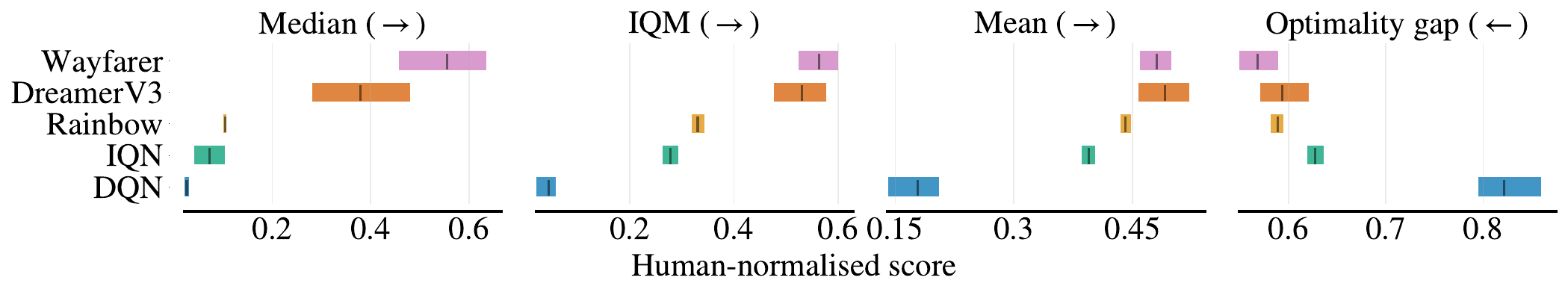}
  \caption{Aggregate metrics on the Atari 2600 challenging set, over 200M frames, with 95\% CIs, using AUC of mean episodic returns over the full training period instead of the mean over last 10\%. Higher mean, median, and IQM and lower optimality gap are better.} %
  \label{fig:challengingresultsauc}
\end{figure}

\paragraph{Area under the curve.}
An alternative metric for reporting results in the Arcade Learning Environment is the area under the learning curve (AUC).
When we use AUC over the 200M frames, IQN and Rainbow flip places in the ranking.
Wayfarer's relative performance improves according to the mean, and it now performs comparably to or better than DreamerV3 on every metric (\autoref{fig:challengingresultsauc}).

\begin{figure}[htpb]
    \centering
    \includegraphics[width=0.62\linewidth]{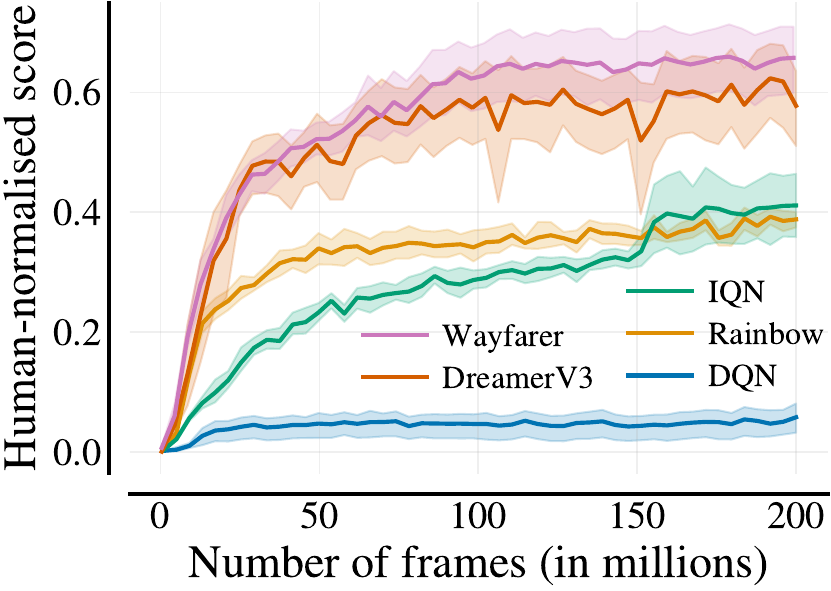}
    \caption{Interquartile mean (IQM) human-normalised scores (HNS) with 95\% bootstrapped CIs.}
    \label{fig:challengingsampleeff}
\end{figure}

\paragraph{Sample efficiency.}
We plot IQM scores as a function of training frames (\autoref{fig:challengingsampleeff}).

\subsection{Per-game training curves: Wayfarer vs. baselines}

\begin{figure}[h]
  \centering
  \includegraphics[width=1.00\linewidth]{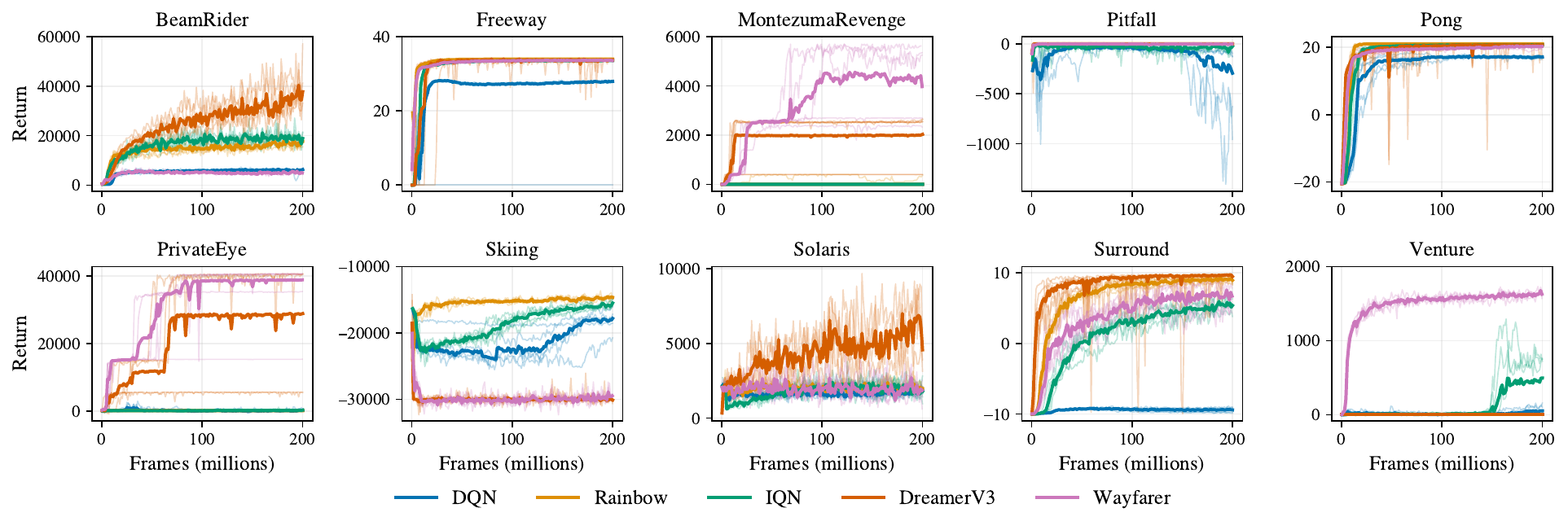}
  \caption{Mean episodic returns per game over 200M frames.
  The individual runs are plotted faintly, with a darker line showing the IQM across seeds (5 per game for all methods except DreamerV3, which has 4--6).}
  \label{fig:pergame}
\end{figure}

\subsection{Per-game training curves: Wayfarer vs. ablations}

\begin{figure}[h]
  \centering
  \includegraphics[width=1.00\linewidth]{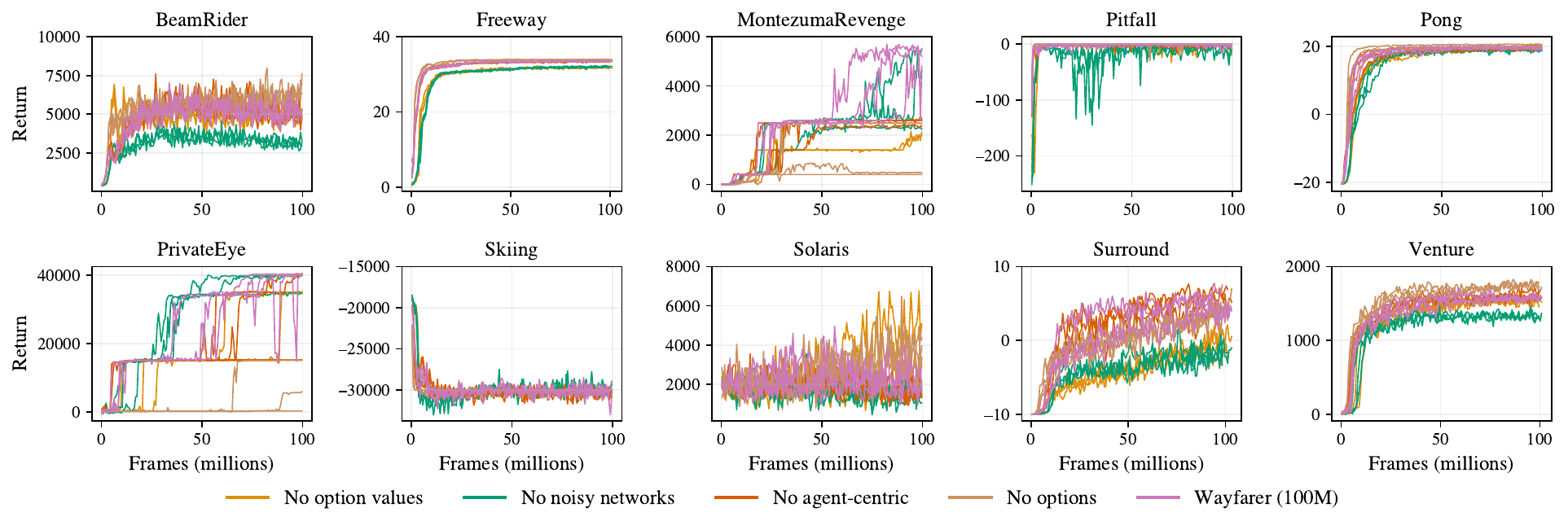}
  \caption{Mean episodic returns per game over 100M frames.
  The individual runs are plotted as separate lines (3 per configuration per game).}
  \label{fig:pergameabla}
\end{figure}

\subsection{Tables of scores}

\begin{longtable}{l | r r | r r r r r}
\caption{Mean episodic returns from the last $10\%$ of training: \textbf{median} across all available seeds (5 per game for all methods except DreamerV3, which has 4--6).} \label{table:challengingresultsmedian} \\
\toprule
Task & Random & Human & DQN & Rainbow & IQN & DreamerV3 & Wayfarer \\
\midrule
\endfirsthead

\toprule
Task & Random & Human & DQN & Rainbow & IQN & DreamerV3 & Wayfarer \\
\midrule
\endhead

\midrule
\multicolumn{8}{r}{\textit{Continued on next page...}} \\
\endfoot

\bottomrule
\endlastfoot

\input{tables/main_results_median_body}
\end{longtable}

\begin{longtable}{l | r r | r r r r r}
\caption{Mean episodic returns from the last $10\%$ of training: \textbf{IQM} across all available seeds (5 per game for all methods except DreamerV3, which has 4--6).} \label{table:challengingresultsiqm} \\
\toprule
Task & Random & Human & DQN & Rainbow & IQN & DreamerV3 & Wayfarer \\
\midrule
\endfirsthead

\toprule
Task & Random & Human & DQN & Rainbow & IQN & DreamerV3 & Wayfarer \\
\midrule
\endhead

\midrule
\multicolumn{8}{r}{\textit{Continued on next page...}} \\
\endfoot

\bottomrule
\endlastfoot

\input{tables/main_results_iqm_body}
\end{longtable}

\begin{longtable}{l | r r | r r r r r}
\caption{Mean episodic returns from the last $10\%$ of training: \textbf{mean} across all available seeds (5 per game for all methods except DreamerV3, which has 4--6).} \label{table:challengingresultsmean} \\
\toprule
Task & Random & Human & DQN & Rainbow & IQN & DreamerV3 & Wayfarer \\
\midrule
\endfirsthead

\toprule
Task & Random & Human & DQN & Rainbow & IQN & DreamerV3 & Wayfarer \\
\midrule
\endhead

\midrule
\multicolumn{8}{r}{\textit{Continued on next page...}} \\
\endfoot

\bottomrule
\endlastfoot

\input{tables/main_results_mean_body}
\end{longtable}

\section{Baselines}
\label{app:baselines}

Wayfarer is a \emph{single-stream} agent: one learner, acting in one instance of the environment, learning online from the experience it generates.
Rainbow and IQN provide established single-stream, model-free baselines whose components underlie Wayfarer.
DreamerV3 \citep{hafner2025mastering} is a state-of-the-art reference beyond the model-free class.
Dreamer learns from imagined experience and collects data from parallel environment instances, so this comparison places Wayfarer's performance in a broader context while differing in architecture and paradigm.

\paragraph{Data source.}
The data for Rainbow, IQN, and DQN come from \emph{DQN Zoo} \citep{quan2020dqn}, a collection of reference implementations published by Google DeepMind.
The data for DreamerV3 come from the original publication \citep{hafner2025mastering}.
The DQN Zoo agents deviate from our training protocol in a few environment settings: the model-free baselines use minimal action sets, no-op starts without sticky actions, and life information.
Wayfarer trains with sticky actions, the full action set, and no access to life information beyond the observations.
These differences may favour the baselines. %

\paragraph{Ablated configurations.}
We selected the hyperparameters for our ablated configurations of Wayfarer as follows.
For \emph{no options}, we tuned the NoisyNet initial noise level on the challenging set.
For \emph{no noisy networks} and \emph{no option values}, the two configurations that use $\epsilon$-greedy, we took the $\epsilon$-decay schedule from \citet{dabney2021temporally-extended}: the parameter follows a linear decay from 1.0 to 0.01 over the first 4M frames and remains constant after that.
We tuned \emph{no agent-centric} on Montezuma's Revenge by evaluating the top 15 Wayfarer configurations without agent-centric regularisation.
In the end, we used the same configuration as Wayfarer (minus the regularisation).
Where applicable, we used the same option termination probability as Wayfarer (\autoref{app:hypers}).

\section{What does the agent learn to represent?}
\label{app:representation}

\begin{figure}[h]
  \centering
  \includegraphics[width=0.27\linewidth, height=0.27\linewidth, keepaspectratio=false]{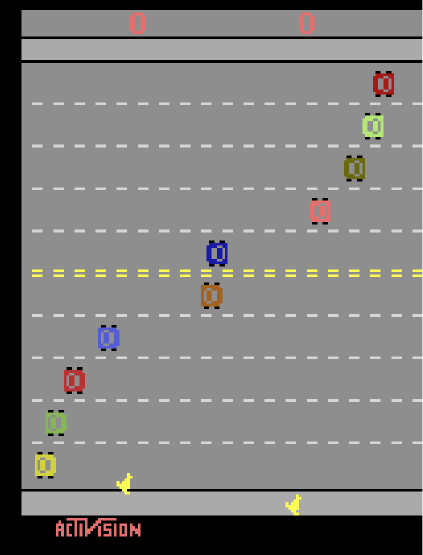}\hspace{0.02\linewidth}
  \includegraphics[width=0.27\linewidth, height=0.27\linewidth]{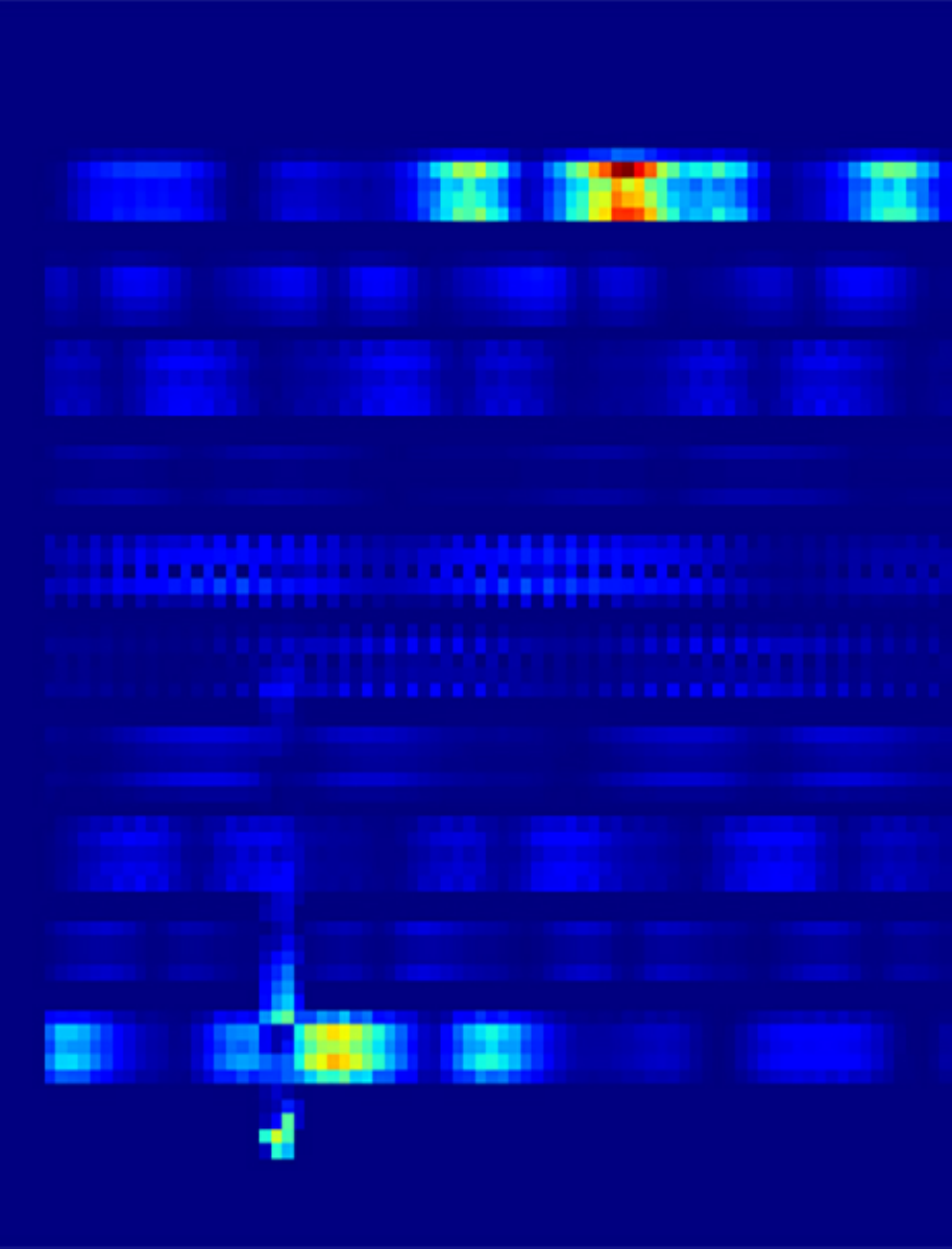}\hspace{0.02\linewidth}
  \includegraphics[width=0.27\linewidth, height=0.27\linewidth]{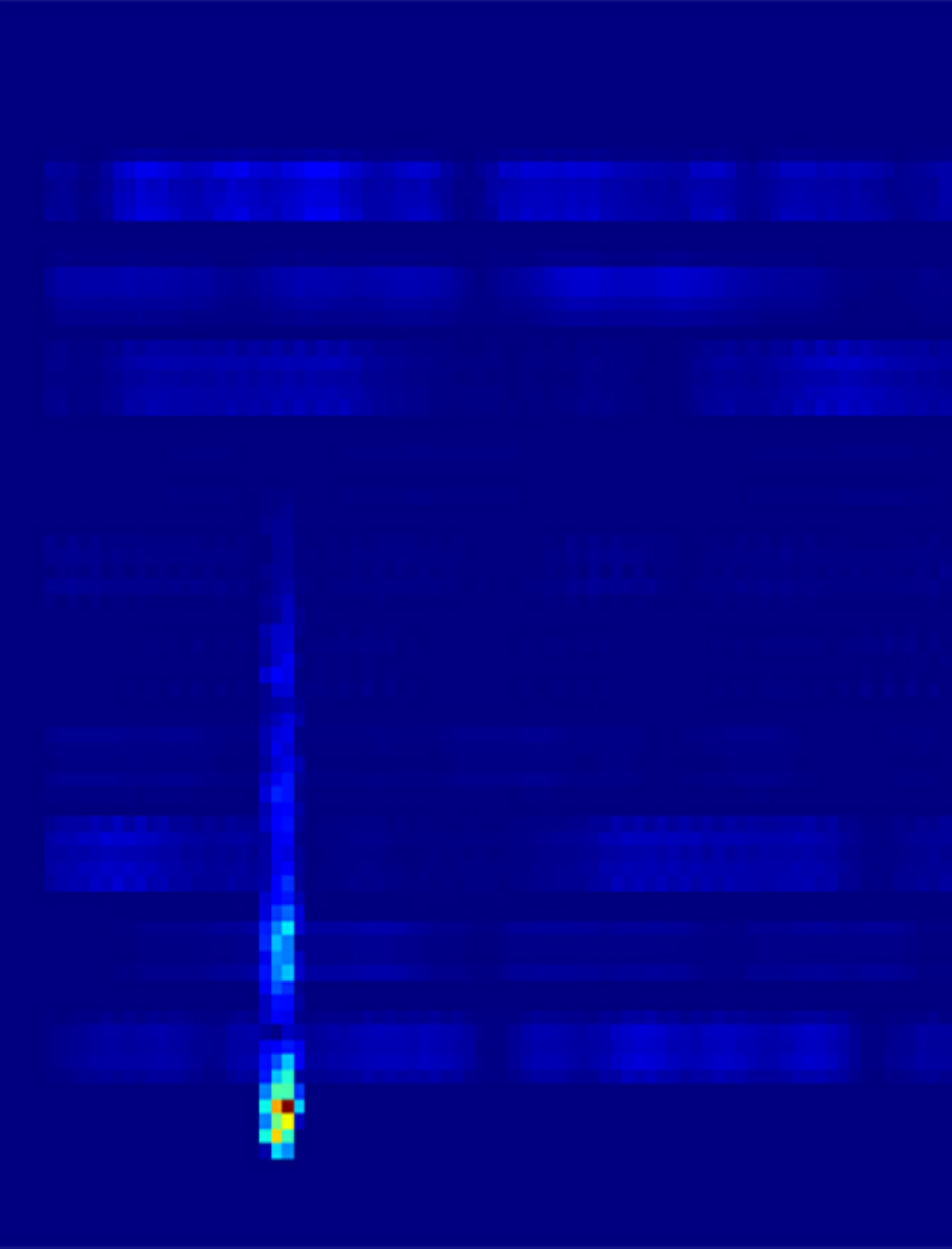}
  \caption{Freeway. \figleft{}. An RGB frame from the game. \figcenter{} The smoothest non-constant eigenfunction tracks exogenous structure---the two slowest cars---whose motion is unaffected by the agent. \figright{} With agent-centric regularisation, the smoothest non-constant eigenfunction tracks only the agent's vertical position, which is the part of the observation under the agent's control.}
  \label{fig:freewayattribution}
\end{figure}

In the game Freeway, the objective is to cross the road as many times as possible, while avoiding cars, within a limited amount of time.
The agent can only move up or down.
\autoref{fig:freewayattribution} visualises the smoothest non-constant eigenfunction learned with Laplacian representation learning directly from observations versus that learned from a contingent-aware feature space.
It shows the problem described in \autoref{sec:repnetwork} in action: without agent-centric regularisation, the two options corresponding to the visualised smoothest non-constant eigenfunction could receive intrinsic rewards driven by changes outside the agent's control (it cannot affect the motion of the cars).
With regularisation, the eigenfunction instead appears to track the agent's vertical position: the agent can only move up or down along the vertical path.

\begin{figure}[h]
  \centering
  \includegraphics[width=0.27\linewidth, height=0.27\linewidth]{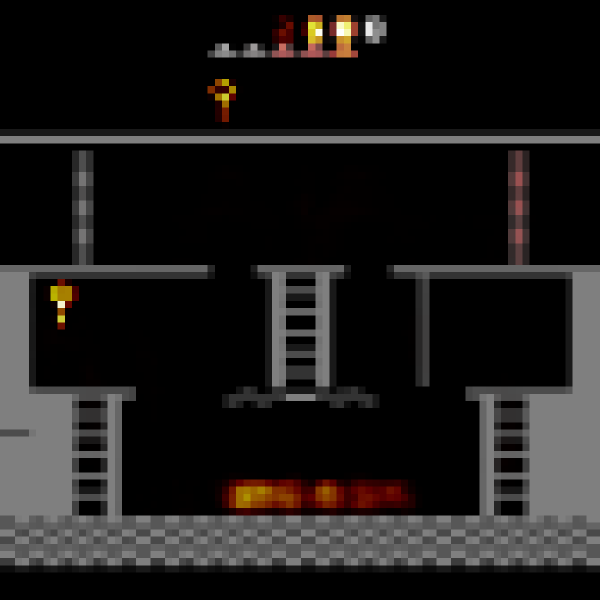}\hspace{0.02\linewidth}
  \includegraphics[width=0.27\linewidth, height=0.27\linewidth]{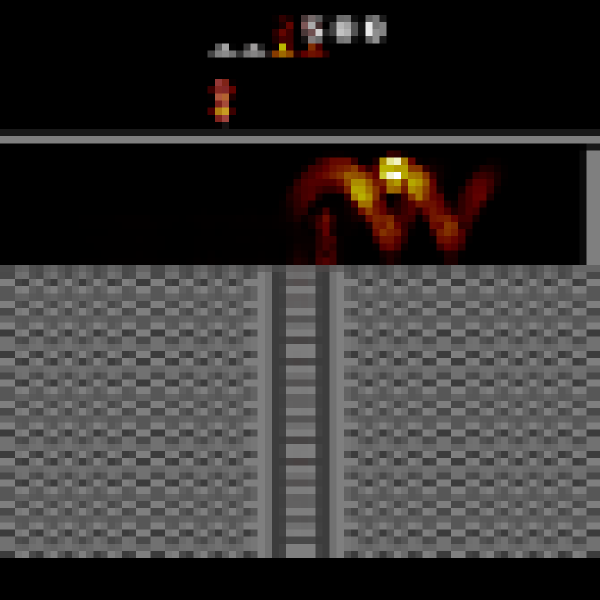}\hspace{0.02\linewidth}
  \includegraphics[width=0.27\linewidth, height=0.27\linewidth]{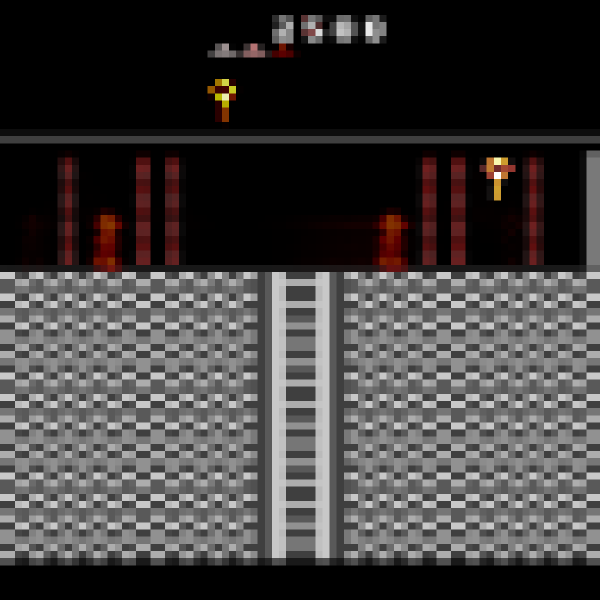}
  \caption{Montezuma's Revenge (rooms 1, 2, and 7). The learned Laplacian representation is sensitive to changes in pixels associated with progress, including keys, doors, skulls, laser gates, and parts of the heads-up display. When these features change, the behaviours that can lead to further progress may also change, making them plausible indicators of transitions between weakly connected regions of the experienced state space.}
  \label{fig:mrattribution}
\end{figure}

\autoref{fig:mrattribution} provides a lens on the representation learning in Montezuma's Revenge.
The mechanics of the game are explained in Appendix~\ref{app:montezuma}, alongside a map of the first level to provide some spatial context (\autoref{fig:mrlevel1}).
We visualise which parts of the agent's observations, in three of the first rooms, the representation is most sensitive to.
The representation appears to track features associated with the environment's \emph{bottlenecks}.
Bottlenecks, by definition, gate transitions between weakly connected regions in an environment \citep{simsek2008skill}.
The smoothest non-constant eigenfunctions tend to spike between such regions [\citealp{machado2023temporal}, Figure~4, p.~11], so options derived from them may traverse such bottlenecks. %

\subsection{Feature attribution}

We analyse the learned Laplacian eigenfunctions using the basis-invariant attribution method of \citet{hilton2024a}.
The method can be viewed as a \emph{sensitivity-weighted analogue of PCA}: it identifies directions of variation in the observations that matter to a given eigenfunction, accounting for both their variance in the input space and the eigenfunction's sensitivity to them.

Given a fixed representation network checkpoint, let $u_i:\RR^m\rightarrow\RR$ denote the $i$-th learned eigenfunction, where $m$ is the dimensionality of a flattened observation $\vec{x}\in\RR^m$.
Let $\D_{\mathrm{attr}}$ denote the dataset of observations used for feature attribution; we write $\E_{\D_{\mathrm{attr}}}$ for an empirical average over these observations.

Define the observation mean $\boldsymbol{\mu}$ and covariance matrix $\mat{\Sigma}$, and the uncentred gradient covariance matrix $\mat{\Gamma}_i$, by
\begin{align*}
  \boldsymbol{\mu} &:= \E_{\D_{\mathrm{attr}}}[\vec{x}]\,, \\
  \mat{\Sigma} &:= \E_{\D_{\mathrm{attr}}}\bigl[
    (\vec{x}-\boldsymbol{\mu})
    (\vec{x}-\boldsymbol{\mu})^\top
  \bigr]\,, \\
  \mat{\Gamma}_i &:= \E_{\D_{\mathrm{attr}}}\bigl[
    \nabla_{\vec{x}}u_i(\vec{x})
    \nabla_{\vec{x}}u_i(\vec{x})^\top
  \bigr]\,,
\end{align*}
where $\nabla_{\vec{x}}$ denotes the gradient with respect to the observation $\vec{x}$.

The method seeks a projection matrix $\mat{P}\in\RR^{m\times m}$ satisfying $\mat{P}^2=\mat{P}$ and $\operatorname{rank}(\mat{P})\leq k$, where $k$ is the maximum number of directions we wish to retain.
Following the derivation in \citet{hilton2024a}, we approximately minimise the surrogate objective
\begin{align*}
  \mathcal{L}_{\mathrm{attr},i}(\mat{P}) := \Tr\!\left(
    (\mI_m-\mat{P})\mat{\Sigma}
    (\mI_m-\mat{P})^\top\mat{\Gamma}_i
  \right),
\end{align*}
where $\mI_m$ is the $m\times m$ identity matrix and $\Tr$ denotes the trace.

The matrix $\mat{\Gamma}_i$ weights reconstruction errors by the sensitivity of $u_i$.
When $\mat{\Gamma}_i=\mI_m$, the objective reduces to the PCA reconstruction objective.

We use the fast approximate algorithm described in Section~4.1 of \citet{hilton2024a}, based on randomised subspace iteration \citep{halko2011finding}.
The algorithm returns a matrix $\mat{P}_A$ whose columns form a basis for the image of $\mat{P}$.
We reshape each column of $\mat{P}_A$ to the observation dimensions to obtain an attribution map.

\paragraph{Applying the method to Atari 2600 games.}
We use observations stored in experience replay buffers saved alongside the network checkpoints.
\autoref{fig:freewayattribution} shows two maps from the game Freeway, each computed separately for the smoothest non-constant eigenfunction from the corresponding training run.
For Montezuma's Revenge, we used a vector-valued extension of the method \citep[Section~4.4]{hilton2024a}.
For an observation $\vec{x}\in\RR^m$, the representation network outputs $\vec{u}=f_\phi\bigl(f_\psi(\vec{x})\bigr)\in\RR^d$, where $d$ is the number of learned eigenfunctions, $f_\psi$ is the shared encoder, and $f_\phi$ is the Laplacian head (\autoref{sec:repnetwork}).
Then, let
\begin{align*}
  \mat{J}_{\vec{u}}(\vec{x})
  := \frac{\partial\vec{u}}{\partial\vec{x}}
  \in\RR^{d\times m}
\end{align*}
denote the Jacobian of this output with respect to the observation.
The vector-valued extension replaces $\mat{\Gamma}_i$ with
\begin{align*}
  \mat{\Gamma} := \E_{\D_{\mathrm{attr}}}\bigl[
    \mat{J}_{\vec{u}}(\vec{x})^\top
    \mat{J}_{\vec{u}}(\vec{x})
  \bigr].
\end{align*}
This gives one set of directions for the Laplacian representation as a whole.
We partitioned the saved observations by room and, for each room, used the resulting subsample as $\D_{\mathrm{attr}}$ to compute a separate $\mat{P}_A$.
One such attribution map is shown in the top-right panel of \autoref{fig:frontpage}.
In \autoref{fig:mrattribution}, the maps are overlaid on room templates formed by taking the pixel-wise median of observations from the corresponding room.

\section{Extended related work}
\label{app:extendedrelated}

Wayfarer builds on representation-driven option discovery, on Laplacian and agent-centric representation learning, and on the broader option-discovery literature.
As such, we discuss how Wayfarer relates to existing approaches in the literature.

\subsection{Learning and using Laplacian-based options online}

Like Wayfarer, Deep Covering Eigenoptions (DCEO) \citep{klissarov2023deep} learns the smoothest eigenfunctions of the Laplacian online, from high-dimensional observations, and derives options from them, instantiating the representation-driven option discovery (ROD) cycle \citep{machado2023temporal,machado2025representation-driven}.
DCEO, however, uses its options only for exploration: when its $\epsilon$-greedy policy explores, the agent follows a randomly chosen option until termination.
Indeed, \citet{klissarov2023deep} explicitly identify learning option values as a direction for future work (p.~9).
Wayfarer's main network instead learns task-reward values over primitive actions and options alike, and chooses among them by their estimated return.
Wayfarer also differs in other ways: it regularises its representation with multi-step inverse dynamics (\autoref{sec:repnetwork}), learns the representation with ALLO \citep{gomez2024proper} rather than GGDO \citep{wang2021towards}, uses a unified skill network to learn the options, and uses IQN \citep{dabney2018implicit} rather than C51 \citep{bellemare2017a} with Rainbow.

Learning option values introduces difficulties of its own.
\citet{kotamreddy2025a} show that, with exploration controlled for, learning eigenoption values can accelerate credit assignment in tabular settings.
However, they struggled to show gains when learning online or with function approximation.
Wayfarer learns option values in exactly this setting: online, with deep neural networks, from high-dimensional observations.
Our ablations separate the effect of learning option values from that of the exploration mechanism, and include a DCEO-style variant that uses the options only for exploration (\autoref{sec:experiments}).

Novel Exploration via Orthogonality (NEO) \citep{theophilou2025novel} also discovers options online and learns their values from task reward, but in tabular domains.
NEO adds visitation counts to a modified Laplacian, so that its eigenvectors, and the option policies defined from them, lead towards novel parts of the state space.
NEO learns option values with SMDP Q-learning over entire option executions.
The algorithm computes the eigenvectors of this modified Laplacian, with option policies defined over the eigenvectors.
Wayfarer instead learns both its representation and its option policies from high-dimensional observations, and its main network updates option values with the same $n$-step target as primitive actions regardless of option duration (\autoref{sec:mainnetwork}).

Hierarchical Novel Eigen Options (HNEO) \citep{vijayan2026addressing} extends NEO to pixel-based domains.
They do this by approximating counts and using this as a reward bonus on top of the rewards given by the Laplacian eigenfunctions.
Like NEO, HNEO learns option values with SMDP updates.
HNEO builds on PQN \citep{gallici2025simplifying} and learns from many parallel instances of the environment, whereas Wayfarer learns from a single stream of experience.
The two methods also regularise Laplacian representation learning differently: HNEO towards novelty, Wayfarer towards agent-centric features.

Earlier work established several parts of Wayfarer's approach to ROD.
\citet{machado2018eigenoption} derived Laplacian-based options from raw pixels, learning a deep successor representation \citep{dayan1993improving} from a fixed dataset collected by a uniformly random policy.
Deep covering options \citep{jinnai2020exploration} scale covering options \citep{jinnai2019discovering} to large state spaces: each option terminates in an under-explored region identified by an approximate Laplacian eigenfunction \citep{wu2019the} and joins the action set of a higher-level policy that learns its value from task reward.
Options are added one at a time, and the method was evaluated quantitatively in low-dimensional state spaces, with qualitative results in Atari.
Wayfarer likewise values its options with task reward, but does not only rely on the smoothest (non-constant) eigenfunction and learns all options simultaneously in one online loop from high-dimensional experience.

Continual curiosity-driven skill acquisition (CCSA) \citep{kompella2017continual} was among the first methods to couple perceptual abstraction and skill acquisition from raw pixels, storing and reusing acquired skills to support further skill discovery.
CCSA learns slow features, which have been shown to be mathematically related to Laplacian eigenmaps \citep{sprekeler2011on}, and intrinsically rewards transitions that produce large changes in their outputs.
CCSA learns each skill by model-based policy iteration over a transition model of abstract states obtained by clustering the outputs of the slow features.
Wayfarer's options likewise seek change in slowly varying features, but each is rewarded for moving one eigenfunction in one direction, and all are discovered without a learned transition model using deep Q-learning.

\subsection{Unsupervised skill discovery for downstream tasks}

Unsupervised skill discovery asks how an agent can acquire useful behaviours without task rewards.
Variational intrinsic control \citep{gregor2016variational}, for example, maximises mutual information between options and their termination states, encouraging behaviours with distinguishable outcomes.
Metric-Aware Abstraction (METRA) \citep{park2024metra} grounds skill discovery in a temporal-distance metric and scales to locomotion from pixels.
\citet{zheng2025can} analyse METRA through the lens of mutual-information skill learning, derive a connection between its representation learning objective and contrastive learning, and introduce Contrastive Successor Features (CSF).
In their hierarchical experiments, METRA and CSF first pre-train skills and then train a controller whose actions are those skills.
Wayfarer does not separate these phases: its task-directed choices shape the experience from which it learns its representation and options, and the changing options in turn shape the behaviours available to it.

\subsection{Spectral and agent-centric representations}

ROD requires a representation that remains useful as the agent's experience changes, yet spectral graph-drawing objectives such as ALLO \citep{gomez2024proper} are typically trained under a fixed, often uniformly random, policy \citep{ahmed2025online}.
\citet{erraqabi2022temporal} observe that the temporally contrastive formulation of the Laplacian assumes uniform access to all states, and pair representation learning with a skill-based covering policy that improves the training distribution.
In Wayfarer, the options that broaden this distribution are those the main network selects for task return (as well as those it explores with) and not a separate covering policy.
\citet{ahmed2025online} study Laplacian representation learning under a changing policy: they introduce the Asymmetric Graph Drawing Objective (AGDO), prove that its online optimisation converges when policy drift is bounded, and find that accuracy suffers when it is not.
Wayfarer uses ALLO, but holds its penalty coefficient fixed and applies a symmetric log transformation to its outputs to stabilise learning from changing experience (\autoref{sec:repnetwork}).
Combining Wayfarer with an objective like AGDO is left for future work.

Spectral structure can also drive exploration without options: the Spectral Bellman Method \citep{nabati2026spectral} learns state-action features aligned with the structure of Bellman updates and explores by Thompson sampling in the learned feature space.
In comparison, Wayfarer uses its Laplacian eigenfunctions to define rewards for temporally extended behaviours.

The source of temporal variation also matters when a representation defines rewards for behaviour.
A timer can change slowly regardless of the agent's actions, providing a predictable feature without providing a useful objective for an option.
\emph{Contingency awareness} \citep{bellemare2012investigating} captures the distinction between changes influenced by the agent and those determined by the environment, and has been used to construct features for value approximation in Atari.
More recent work formalises the problem of extracting endogenous state information in the presence of exogenous distractors, developing methods based on multistep inverse dynamics prediction \citep{efroni2022provably,lamb2023guaranteed,islam2023principled}.
Single-step inverse dynamics has also been used to learn the feature space in which prediction-error intrinsic rewards are computed \citep{pathak2017curiosity}, and to focus novelty on controllable aspects of Atari observations \citep{badia2020never,badia2020agent57}.
Wayfarer regularises the encoder shared with its Laplacian head by multi-step inverse-dynamics prediction, which encourages the representation network to encode information from which the agent's actions can be inferred, thereby shaping the eigenfunctions that define option rewards.
The regulariser supplies an inductive bias and our experiments examine its effects (\autoref{sec:experiments}, \autoref{app:representation}).

Empowerment offers another link between skill learning and control-relevant representations: \citet{bastankhah2026learning} show that agents maximising empowerment induce representations invariant to control-irrelevant features.
Wayfarer pursues a related aim through inverse-dynamics regularisation of a spectral representation.

\subsection{Abstractions from task rewards and subgoals}

Other methods derive their option-discovery objectives from task rewards or from designer-specified subgoals.
The option-critic architecture \citep{bacon2017the} learns option policies and termination functions directly from task reward.
FeUdal Networks (FuN) \citep{vezhnevets2017feudal} train a manager to set goals in a learned latent space and a worker to pursue them.
The worker receives an intrinsic reward for following the goals, but the manager learns from task reward which goals to set.
In both methods, task reward determines which behaviours are learned, but where non-zero rewards are sparse such methods grounded in reward have no signal to build on.
Wayfarer's representation-derived rewards, in contrast, shape its options before the agent encounters informative task rewards.

Hierarchical DQN (h-DQN) \citep{kulkarni2016hierarchical}, instead of using task reward, learns lower-level policies for designer-given subgoals, as well as a higher-level policy that selects among them.
Similarly, earlier intrinsically motivated RL organised skill acquisition around designer-specified ``salient'' events \citep{singh2004intrinsically}.
These methods show how to learn useful behaviours once their objectives are given, whereas Wayfarer derives those objectives from interaction with the environment.

Task reward can also enter through the representation itself.
\citet{tse2025reward-aware} study the default representation, an analogue of the successor representation that also accounts for the problem's reward dynamics, and show that it gives rise to reward-aware behaviour in option discovery.
Wayfarer keeps the representation reward-free: task reward enters only through the main network, which values the options and, through its choices, shapes the experience from which they are learned.

\citet{bagaria2026going} use feature attribution to identify a small subset of relevant features for each subgoal, and learn options that reach the resulting abstract subgoals.
Wayfarer uses feature attribution only to analyse its learned eigenfunctions (\autoref{app:representation}), and feature attribution plays no part in its option rewards or termination.

Skill-driven neurosymbolic state abstractions \citep{ahmetoglu2025skill-driven} reverse the dependence at the core of ROD: rather than deriving options from a representation, they construct abstract states compatible with a given set of abstract actions.
It is an interesting complementary approach to Wayfarer; we leave thoughts about fusing the approaches for future work.

\section{Normalisation and effective step size}
\label{app:nap}

\begin{wrapfigure}{r}{0.32\textwidth}
  \centering
  \vspace{-\intextsep}
  \includegraphics[width=\linewidth]{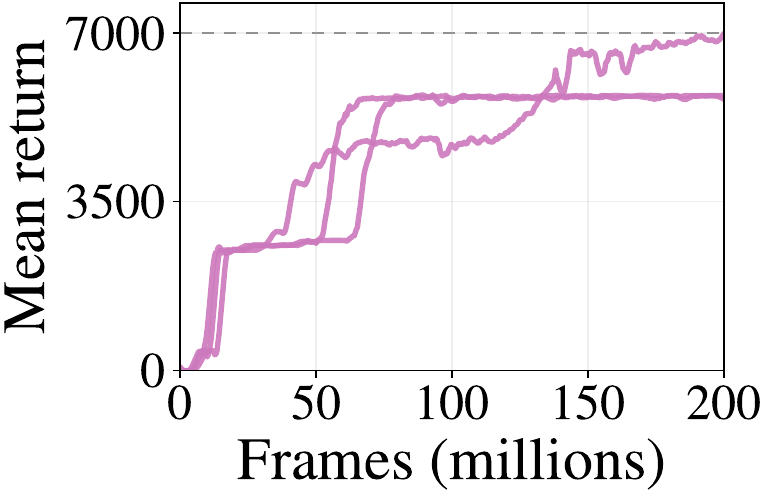}
  \captionsetup{font=small}
  \caption{Mean episodic return in Montezuma's Revenge with NaP and a constant effective step size.}
  \label{fig:mr7k}
\end{wrapfigure}
We observed that Wayfarer can plateau early in games requiring continual exploration, and we suspect that implicit step size decay contributes to these plateaus.
With LayerNorm \citep{ba2016layer}, used to stabilise deep Q-learning \citep{gallici2025simplifying}, growing weight norms implicitly reduce the effective step size thus slowing adaptation to new experience \citep[pp.~4--5]{lyle2024normalization}.
We experimented with Normalize-and-Project (NaP) \citep{lyle2024normalization}, a simple modification that fixes weight norms in normalised layers to prevent this implicit decay.
With the effective step size held constant, Wayfarer achieved a mean episodic return of over 7,000 on Montezuma's Revenge within the 200M frame budget (\autoref{fig:mr7k}).$^\triangle$
This result motivates further work on sustaining plasticity through mechanisms such as NaP.

\end{document}

%% file: math_commands.tex
\usepackage{amsmath,amsfonts}
\usepackage{eucal}
\usepackage{bm}
\usepackage{ stmaryrd }   %

\newcommand{\figleft}{{\em (Left)}}
\newcommand{\figcenter}{{\em (Center)}}
\newcommand{\figright}{{\em (Right)}}

\def\eqref#1{equation~\ref{#1}}

\def\1{\bm{1}}

\def\eps{{\epsilon}}

\def\rva{{\mathbf{a}}}

\def\rvu{{\mathbf{i}}}

\def\rvu{{\mathbf{u}}}

\def\mI{{\bm{I}}}

\DeclareMathAlphabet{\mathsfit}{\encodingdefault}{\sfdefault}{m}{sl}
\SetMathAlphabet{\mathsfit}{bold}{\encodingdefault}{\sfdefault}{bx}{n}

\def\gA{{\mathcal{A}}}

\def\gO{{\mathcal{O}}}

\def\gR{{\mathcal{R}}}
\def\gS{{\mathcal{S}}}

\def\gX{{\mathcal{X}}}

\newcommand{\E}{\mathbb{E}}

\newcommand{\softmax}{\mathrm{softmax}}

\DeclareMathOperator*{\argmax}{arg\,max}

\DeclareMathOperator{\Tr}{Tr}

\newcommand{\cS}{\mathcal{S}}

\newcommand{\D}{\mathcal{D}}
\newcommand{\RR}{\mathbb{R}}

\renewcommand{\vec}[1]{\mathbf{#1}}

\newcommand{\mat}[1]{\mathbf{#1}}

\newcommand{\ip}[1]{\langle #1 \rangle}

\newcommand{\id}{\mat{I}}
\newcommand{\pp}{\mat{P}_{\!\pi}}

\newcommand{\la}{\mat{L}}

\newcommand{\nS}{|\cS|}
\newcommand{\sg}[1]{\llbracket #1 \rrbracket}
\newcommand{\dual}{\boldsymbol{\beta}}

\newcommand{\loss}{\mathcal{L}}

\newcommand{\hsq}{h}  %
\newcommand{\hinv}{h^{-1}}  %

\providecommand{\proofname}{Proof}

\providecommand{\proofsketchname}{Proof Sketch}

%% file: wf_networks.tex
\colorlet{conv}{cyan!15}
\colorlet{mlp}{green!15}
\colorlet{noisy}{red!15}
\colorlet{auxiliary}{black!50}

\begin{tikzpicture}[
  >={Stealth[length=3mm, width=2.3mm]},
  every node/.style={text=black},
  flow/.style={->, line width=1.2pt, draw=black},
  auxflow/.style={->, line width=1.2pt, draw=auxiliary, dashed},
  coupling/.style={->, line width=1.0pt, draw=black!50, densely dotted},
  screen/.style={draw=black, line width=1.1pt, rounded corners=2pt,
                 fill=black!80, minimum width=1.9cm, minimum height=1.5cm},
  ghostscreen/.style={draw=auxiliary, dashed, line width=1.1pt, rounded corners=2pt,
                 fill=black!12, minimum width=1.9cm, minimum height=1.5cm},
  backbone/.style={trapezium, trapezium left angle=65, trapezium right angle=65,
                 draw=black, line width=1.2pt, fill=conv,
                 minimum width=2.3cm, minimum height=1.25cm, inner sep=2pt},
  headblock/.style={rectangle, rounded corners=2pt, draw=black, line width=1.2pt,
                 align=center, minimum height=0.95cm},
  stackplate/.style={rounded corners=2pt, draw=black, line width=1.2pt, fill=mlp},
  merge/.style={circle, draw=black, line width=1.2pt, fill=white,
                 inner sep=0.5pt, minimum size=0.45cm},
  outtext/.style={font=\large},
  prediction/.style={font=\large, align=center, inner xsep=4pt, inner ysep=5pt},
  coltitle/.style={font=\large},
  couplab/.style={font=\bfseries\itshape, text=black!55, align=center},
]

\newcommand{\sprites}[1]{%
  \fill[red!80!black!70]   ($(#1)+(-0.44,0.38)$)  rectangle ++(0.17,0.17);
  \fill[yellow!85!black]   ($(#1)+( 0.26,-0.02)$) rectangle ++(0.17,0.17);
  \fill[cyan!65]           ($(#1)+(-0.16,-0.48)$) rectangle ++(0.38,0.11);
  \fill[green!65!black!80] ($(#1)+( 0.40,0.42)$)  rectangle ++(0.11,0.11);
}

\begin{scope}[local bounding box=panelA]
\node[ghostscreen] (g1) at (0.66,0.44) {};
\node[screen] (s1) at (0,0) {};
\sprites{s1.center}
\node[anchor=north, font=\bfseries\Large] at ($(s1.west)+(-0.4,0)$) {$x_t$};
\node[anchor=west, font=\bfseries\Large, text=auxiliary]
     at ($(g1.east)+(0.04,0.14)$) {$x_{t+\Delta}$};

\node[backbone] (c1) at (0,2.2) {\huge $\psi$};
\draw[flow] (s1.north) -- (c1.south);
\draw[auxflow] (g1.north) -- (g1.north |- c1.south);

\node[headblock, draw=auxiliary, fill=mlp!55, text=auxiliary,
      minimum width=1.5cm] (chi) at (-3.15,2.2) {\LARGE $\chi$};
\draw[auxflow] (c1.170) -- ($(chi.east)+(0,0.20)$);
\draw[flow, draw=auxiliary] (c1.190) -- ($(chi.east)+(0,-0.20)$);
\node[prediction, text=auxiliary] (chival) at (-3.15,3.72)
     {Inv. dyn. over $\mathcal{A}$\\(regulariser)};
\draw[flow, draw=auxiliary] (chi.north) -- (chival.south);

\node[headblock, fill=mlp, minimum width=1.6cm] (phi) at (0,4.5) {\LARGE $\phi$};
\draw[flow] (c1.north) -- (phi.south);
\node[outtext] (phival) at (0,6.0) {Laplacian eigenfunctions};
\draw[flow] (phi.north) -- (phival.south);
\end{scope}

\begin{scope}[local bounding box=panelB]
\node[screen] (s2) at (6.85,0) {};
\sprites{s2.center}
\node[anchor=north, font=\bfseries\Large] at ($(s2.west)+(-0.4,0)$) {$x_t$};

\node[backbone] (c2) at (6.85,2.2) {\huge $\zeta$};
\draw[flow] (s2.north) -- (c2.south);

\node[merge] (m2) at (6.85,3.4) {\Large $\boldsymbol{\cdot}$};
\draw[flow] (c2.north) -- (m2.south);

\node[font=\bfseries\Large] (t2) at (9.8,0.95) {$\tau\sim U[0,1]$};
\node[headblock, fill=mlp] (qe2) at (9.8,2.3)
     {\large quantile\\[-1pt]\large emb.};
\draw[flow] (t2) -- (qe2);
\draw[flow] (qe2.north) to[out=90,in=0] (m2.east);

\node[headblock, fill=mlp, minimum width=1.6cm] (oh) at (6.85,4.5)
     {\LARGE $\omega_{1:K}$};
\begin{scope}[on background layer]
  \draw[stackplate] ($(oh.south west)+(0.30,0.30)$) rectangle ($(oh.north east)+(0.30,0.30)$);
  \draw[stackplate] ($(oh.south west)+(0.15,0.15)$) rectangle ($(oh.north east)+(0.15,0.15)$);
\end{scope}
\draw[flow] (m2.north) -- (oh.south);

\node[outtext] (ohindex) at (6.85,6.0)
{Returns (of the intr. reward) over $\mathcal{A}$};
\draw[flow] ($(oh.north)+(0,0.30)$) -- (ohindex.south);
\end{scope}

\begin{scope}[local bounding box=panelC]
\node[screen] (s3) at (13.7,0) {};
\sprites{s3.center}
\node[anchor=north, font=\bfseries\Large] at ($(s3.west)+(-0.4,0)$) {$x_t$};

\node[backbone] (c3) at (13.7,2.2) {\huge $\theta$};
\draw[flow] (s3.north) -- (c3.south);

\node[merge] (m3) at (13.7,3.4) {\Large $\boldsymbol{\cdot}$};
\draw[flow] (c3.north) -- (m3.south);

\node[font=\bfseries\Large] (t3) at (16.65,0.95) {$\tau\sim U[0,1]$};
\node[headblock, fill=mlp] (qe3) at (16.65,2.3)
     {\large quantile\\[-1pt]\large emb.};
\draw[flow] (t3) -- (qe3);
\draw[flow] (qe3.north) to[out=90,in=0] (m3.east);

\node[headblock, fill=noisy, minimum width=1.6cm] (mh) at (13.7,4.5)
     {\LARGE $\theta$};
\draw[flow] (m3.north) -- (mh.south);

\node[outtext] (mhval) at (13.7,6.0) {Returns over $\mathcal{O}$ ($\mathcal{A}\subset\mathcal{O}$)};
\draw[flow] (mh.north) -- (mhval.south);
\end{scope}

\node[coltitle, anchor=north] at ($(s1 |- panelA.south)+(0,-0.3)$)
     {\textbf{(A)} Representation\ifstandalone\else: \autoref{sec:repnetwork}\fi};
\node[coltitle, anchor=north] at ($(s2 |- panelB.south)+(0,-0.3)$)
     {\textbf{(B)} Skills\ifstandalone\else: \autoref{sec:optionnetwork}\fi};
\node[coltitle, anchor=north] at ($(s3 |- panelC.south)+(0,-0.3)$)
     {\textbf{(C)} Task\ifstandalone\else: \autoref{sec:mainnetwork}\fi};

\node[couplab] (reprrole) at (3.5,3.75)
     {intrinsic rewards\\for option discovery};
\draw[coupling] (phival.east) to[out=0,in=180] (reprrole.west);
\node[couplab] (optionrole) at (10.5,4.70)
     {options augment\\the main action set};
\draw[coupling] ($(oh.east)+(0.30,0.15)$)
     to[out=0,in=180] (optionrole.west);

\end{tikzpicture}

%% file: tables/main_results_median_body.tex
Beam Rider & 363.9 & 16926.5 & 6044.7 & 17143.4 & 18865.5 & 33863.5 & 5168.2 \\
Freeway & 0.0 & 29.6 & 27.9 & 34.0 & 33.6 & 33.5 & 33.7 \\
Montezuma & 0.0 & 4753.3 & 0.0 & 0.0 & 0.0 & 2524.0 & 5107.9 \\
Pitfall & -229.4 & 6463.7 & -86.2 & 0.0 & -41.0 & -0.1 & -2.5 \\
Pong & -20.7 & 14.6 & 17.2 & 21.0 & 20.6 & 20.5 & 20.1 \\
Private Eye & 24.9 & 69571.3 & 121.8 & 100.0 & 171.8 & 39990.2 & 40468.4 \\
Skiing & -17098.1 & -4336.9 & -17933.7 & -14733.7 & -16085.0 & -29923.6 & -29730.2 \\
Solaris & 1236.3 & 12326.7 & 1734.3 & 2111.1 & 1867.4 & 5858.7 & 1869.9 \\
Surround & -10.0 & 6.5 & -9.2 & 9.1 & 5.3 & 9.5 & 7.1 \\
Venture & 0.0 & 1187.5 & 31.1 & 0.0 & 692.2 & 0.0 & 1625.2

%% file: tables/main_results_iqm_body.tex
Beam Rider & 363.9 & 16926.5 & 6059.9 & 17023.1 & 18890.5 & 34681.6 & 5086.0 \\
Freeway & 0.0 & 29.6 & 27.9 & 34.0 & 33.6 & 33.5 & 33.7 \\
Montezuma & 0.0 & 4753.3 & 0.0 & 0.0 & 0.0 & 1998.2 & 4326.7 \\
Pitfall & -229.4 & 6463.7 & -245.6 & 0.0 & -41.7 & -0.1 & -2.9 \\
Pong & -20.7 & 14.6 & 17.1 & 21.0 & 20.6 & 20.5 & 20.2 \\
Private Eye & 24.9 & 69571.3 & 119.4 & 100.0 & 171.2 & 28538.7 & 38749.9 \\
Skiing & -17098.1 & -4336.9 & -18060.3 & -14708.0 & -15905.6 & -29957.3 & -29723.9 \\
Solaris & 1236.3 & 12326.7 & 1753.6 & 2091.4 & 1931.6 & 5858.7 & 1933.6 \\
Surround & -10.0 & 6.5 & -9.4 & 9.0 & 5.4 & 9.5 & 6.6 \\
Venture & 0.0 & 1187.5 & 48.4 & 0.0 & 471.8 & 0.0 & 1622.5

%% file: tables/main_results_mean_body.tex
Beam Rider & 363.9 & 16926.5 & 5882.4 & 16930.2 & 18829.5 & 35640.4 & 5077.1 \\
Freeway & 0.0 & 29.6 & 22.4 & 34.0 & 33.6 & 33.4 & 33.7 \\
Montezuma & 0.0 & 4753.3 & 0.0 & 39.7 & 0.0 & 1828.8 & 4186.9 \\
Pitfall & -229.4 & 6463.7 & -338.1 & 0.0 & -41.8 & -0.1 & -3.1 \\
Pong & -20.7 & 14.6 & 17.0 & 21.0 & 20.6 & 20.4 & 20.1 \\
Private Eye & 24.9 & 69571.3 & 132.2 & 159.9 & 179.6 & 26287.2 & 34418.4 \\
Skiing & -17098.1 & -4336.9 & -18482.4 & -14730.3 & -15882.9 & -29960.7 & -29708.8 \\
Solaris & 1236.3 & 12326.7 & 1745.9 & 2055.0 & 1933.8 & 5829.3 & 1935.4 \\
Surround & -10.0 & 6.5 & -9.4 & 9.0 & 5.4 & 9.5 & 6.7 \\
Venture & 0.0 & 1187.5 & 50.4 & 0.0 & 435.3 & 0.0 & 1618.9